\documentclass[times, review, 10pt]{elsarticle}

\usepackage{amssymb}
\usepackage{amsmath}
\usepackage{amsthm}
\usepackage{bm}
\usepackage{multirow}
\usepackage{graphics}
\usepackage{diagbox}
\usepackage{bbm}
\usepackage{caption}
\usepackage{xcolor}
\usepackage{dsfont}
\usepackage{subfigure}
\usepackage{makecell}
\usepackage{algorithm}
\usepackage{algorithmic}
\usepackage{textcomp}
\usepackage{verbatim} 
\usepackage[algo2e,linesnumbered,lined,boxed,commentsnumbered,ruled]{algorithm2e}

\usepackage{hyperref}
 
\usepackage{cleveref}
\Crefname{figure}{Fig}{Figures}

\journal{Pattern Recognition}

\begin{document}
\begin{frontmatter}

\title{{SAEN-BGS: Energy-Efficient Spiking AutoEncoder Network for Background Subtraction}}

\author[scut]{Zhixuan Zhang}

\author[szu]{ 
Xiao Peng Li}

\author[scut]{Qi Liu,\textit{~{Senior Member,~IEEE}}}
\affiliation[scut]{organization={School of Future Technology},
            addressline={South China University of Technology},
            city={Guangzhou},
            postcode={511442},
            state={Guangdong},
            country={China}}
 
\affiliation[szu]{organization={State Key Laboratory of Radio Frequency Heterogeneous Integration},
            addressline={Shenzhen University},
            city={Shenzhen},
            postcode={518060},
            state={Guangdong},
            country={China}}

\begin{abstract}
Background subtraction (BGS) is utilized to detect moving objects in a video and is commonly employed at the onset of object tracking and human recognition processes. Nevertheless, existing BGS techniques utilizing deep learning still encounter challenges with various background noises in videos, including variations in lighting, shifts in camera angles, and disturbances like air turbulence or swaying trees. To address this problem, we design a spiking autoencoder network, termed SAEN-BGS, based on noise resilience and time-sequence sensitivity of spiking neural networks (SNNs) to enhance the separation of foreground and background. To eliminate unnecessary background noise and preserve the important foreground elements, we begin by creating the continuous spiking conv-and-dconv block, which serves as the fundamental building block for the decoder in SAEN-BGS. Moreover,  in striving for enhanced energy efficiency, we introduce a novel self-distillation spiking supervised learning method grounded in ANN-to-SNN frameworks, resulting in decreased power consumption. In extensive experiments conducted on CDnet-2014 and DAVIS-2016 datasets, our approach demonstrates superior segmentation performance relative to other baseline methods, even when challenged by complex scenarios with dynamic backgrounds.
\end{abstract}

\end{frontmatter}



\section{Introduction}
\label{Introduction}
Background subtraction (BGS) is used to differentiate between moving objects in the foreground and the background in videos captured from real-life settings \cite{9541391,9222365, 9262966, PR1, PR2, PR3}.
It has a wide variety of uses, such as urban traffic detection \cite{Cheung2005}, long-term video monitoring \cite{seo2014recursive}, optical motion capture \cite{CHEN201435}, and social signal processing \cite{VINCIARELLI20091743}. 

Traditional BGS methods concentrate on improving performance in two particular domains. One potential approach is to develop feature representations that possess increased resilience, such as texture features \cite{text_feature}, color features \cite{color_feature} and edge features \cite{edge_feature}. The second option aims to generate background models with greater accuracy, such as Kernel density estimate (KDE) \cite{KDE}, mixture of Gaussians (MoG) \cite{MoG}, subspace learning models \cite{ 7076585}, and principal component analysis (PCA) \cite{7076585}. Nevertheless, they, referred to as pixel-level BGS approaches, rely exclusively on manually created features to classify individual pixels as either foreground or background, and are highly sensitive to changes in the environment, such as differences in lighting and weather conditions. 

Several BGS methods that rely on data-driven deep learning techniques demonstrate improved performance by incorporating convolutional neural networks (CNNs). In general, they can be classified as unsupervised BGS and supervised BGS. While many unsupervised BGS methods, such as SemanticBGS \cite{SemanticBGS}, enhance their performance by integrating semantic segmentation models with traditional BGS algorithms. However, they are erroneously perceived as the foremost subjects in dim lighting or shadows, regardless of alterations in the backdrop. In supervised BGS methods like those outlined in \cite{Braham-CNN-BGS, Interactive-DNN-BGS}, they typically entail training on the first sequence of frames in test videos, resulting in acceptable outcomes yet unresolved of the constraints of unsupervised BGS methods. To address that, various methods utilizing an autoencoder (AE) for nonlinear dimensionality reduction, such as those mentioned in \cite{avss1, 3dfr, FGSegNet, Tezcan2020, Tezcan2021}, have been investigated for reconstructing backgrounds to enhance the segmentation of foregrounds. During this process, they primarily depend on extracting a low-dimensional feature with minimal noise in the encoder, which is subsequently utilized to accomplish the downstream task of background reconstruction through supervised or unsupervised methods in the decoder. Although they can achieve superior segmentation on static backgrounds, there is still progress to be made for dynamic backgrounds. Intuitively, these AE-based BGS methods may inadvertently overlook the necessity of being robust against background noise closely related to the temporal information between frames. Conversely, the decrease in power usage by the trained autoencoder during the inference phase will become increasingly important as the size and complexity of the autoencoder used in background subtraction grow. Practically, an advanced BGS technology utilizing neural networks is expected to provide exceptional performance and energy efficiency during its inference stage.
 
In this paper, we introduce a spiking autoencoder network for background subtraction (SAEN-BGS) to address the issues mentioned above, inspired by the natural noise resistance and sensitivity to temporal sequences of SNNs. To enhance the prominence of the foreground elements and reduce background interference effectively, we develop a continuous spiking conv-and-dconv block as a fundamental building block for constructing the decoder in SAEN-BGS. Afterward, to implement an energy-efficient inference, we further present a self-distillation spiking supervised learning algorithm based on ANN-to-SNN frameworks. Finally, extensive experiments on CDnet-2014 and DAVIS-2016 datasets demonstrate the superiority of the proposed method over baselines. 

Our main contributions are summarized as follows:
\begin{itemize} 
\item To address the background noise stemming from inadequate temporal information extraction, a spiking autoencoder network (SAEN-BGS) is developed using the noise resilience and time-sequence sensitivity of spiking neural networks.
\item This is the first instance of solving background subtraction from a spike-based perspective, where a continuous spiking convolutional and deconvolutional (conv-and-dconv) block is employed to enhance foreground features and diminish background noise within the decoder.
\item To achieve energy efficiency, a novel self-distillation spiking supervised learning method is proposed within the ANN-to-SNN framework.
\item The empirical evaluations on CDnet-2014 and DAVIS-2016 demonstrate the superiority of the proposed method. Remarkably, even in challenging dynamic backgrounds, our method still outperforms other baselines.
\end{itemize}

The remainder of the paper is organized as follows. 
We review some related works in Section 2.
In Section 3, the proposed neural networks are introduced. 
In Section 4, experimental results on real-world datasets demonstrate that the suggested algorithms outperform the SOTA methods. 
Finally, conclusions are drawn in Section 5.

\section{Related Work}
\subsection{{Background Subtraction Method}}
BGS approaches, which aim to distinguish between foreground and background in videos, can be broadly classified as unsupervised or supervised techniques. Unsupervised BGS methods focus on creating a background model to differentiate between foreground and background. For instance, Stauffer \emph{et al.} \cite{GMM1} propose an online Gaussian Mixture Model (GMM) to update parameters pixel-wise and use multiple Gaussian distributions to represent pixel color distributions. Elgammal \emph{et al.} \cite{GMM3} improve upon this by utilizing Kernel Density Estimation (KDE), a non-parametric method, to describe pixel intensity distributions. On the other hand, Barnich \emph{et al.} \cite{ViBe} present a model that utilizes distances to classify pixels through the capture of local spatial features to establish the background model. St-Charles \emph{et al.} \cite{PAWCS} create PAWCS, a word-based approach that combines color and texture characteristics by treating pixels as background words and continually adjusting their trustworthiness according to persistence. Then Braham \emph{et al.} \cite{SemanticBGS} design a post-processing procedure for the background subtraction algorithm using semantic segmentation prediction. SuBSENSE \cite{subsense} improves detection accuracy through the inclusion of color features and pixel-level feedback. More recently, Isik \emph{et al.} \cite{SWCD} introduce SWCD, a pixel-wise sliding-window technique that updates the background model with a dynamic control system, while Lee \emph{et al.} \cite{WisenetMD} devise a multi-step algorithm to decrease false positives in dynamic backgrounds. What is more, An \emph{et al.} \cite{zbs} develop an unsupervised background subtraction approach which constructs open-vocabulary instance-level background models through zero-shot object detection, and then identifies foreground by comparing new frame detections with these models.

Conversely, supervised BGS algorithms adjust neural network parameters via reducing the differences between labels and training frames using loss function optimization. Braham \emph{et al.} \cite{Braham-CNN-BGS} pioneer the use of CNNs in BGS with ConvNets, achieving impressive results on the CDnet-2014 dataset, despite employing a costly patch-based training approach. Wang \emph{et al.} \cite{Interactive-DNN-BGS} propose a multiscale CNN with a cascading structure, which analyzes individual frames without background frames but requires 200 test video frames for training. Similarly, Babaee \emph{et al.} \cite{DeepBS} develop a video-group-optimized CNN for processing all videos. Sakkos \emph{et al.} \cite{3D-CNN-BGS} introduce a 3D CNN designed to capture both temporal and color information. However, these techniques rely on utilizing partial frames from test videos during training, which could limit their effectiveness with previously unseen videos.

\subsection{{Autoencoder-based BGS Method}}
Recent research has integrated nonlinear dimensionality reduction techniques using AE to improve the differentiation between foreground and background components, enabling the extraction of enhanced feature data through an iterative optimization approach. Lim \emph{et al.} \cite{avss1} propose an AE for background subtraction, where the encoder is responsible for extracting feature vectors, and the decoder converts these vectors into segmentation maps by utilizing the current frame, a previous frame, and a background model as input.
In \cite{3dfr}, Mandal \emph{et al.} introduce 3DFR, which involves three feature reduction networks and utilizes 50 past frames to aid in the current frame's learning process. In addition, Lim \emph{et al.} design two robust AEs, known as FgSegNet\_M and FgSegNet\_S \cite{FGSegNet}, which produce multi-scale features in the encoder section. Note that these techniques necessitate a labeled frame from the test video for the purpose of training. Then Tezcan \emph{et al.} propose a CNN-based on UNET architecture called BSUV-Net \cite{Tezcan2020}, which takes the current frame and two background frames as input. At the same time, BSUV-Net 2.0 \cite{Tezcan2021} and Fast BSUV-Net 2.0 \cite{Tezcan2021} are introduced and achieve similar levels of effectiveness.

\begin{figure}[]
\centering
\centerline
{\includegraphics[width = 7cm]{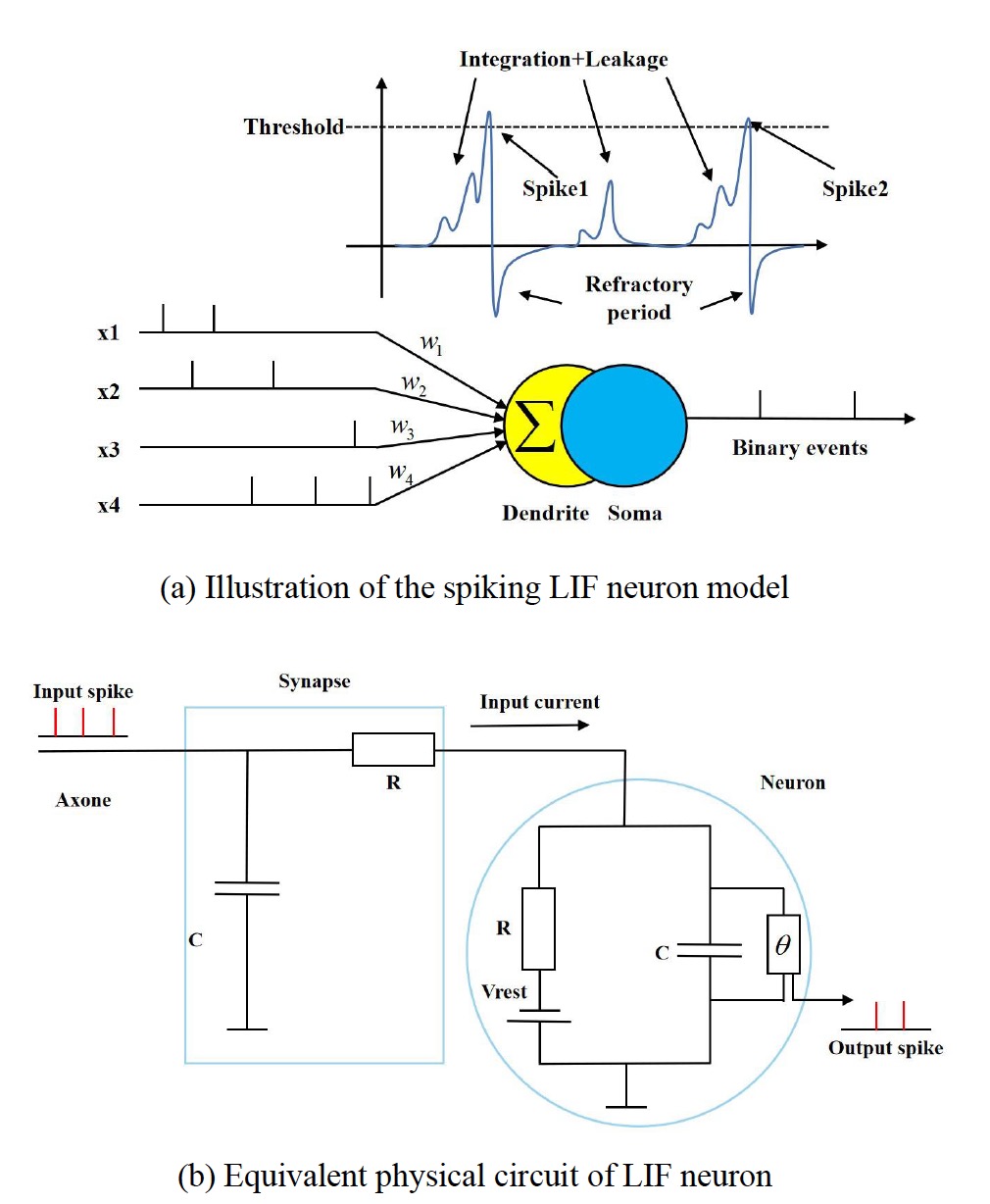}}
\caption{Brain-like leaky integrate-and-fire (LIF) model corresponds to its physical circuit. The soma constantly receives integrated input from dendrites. When the sum of total input exceeds a certain threshold $\vartheta$, an output spike is generated and then delivered to other neurons by the axon. After firing a spike, the membrane potential is reset to $V_{rest}$.}
\label{LIF}
\end{figure}

\subsection{{Low-power Computational Framework}}
Recently, as the parameters and operators of neural networks have grown, the research community has become increasingly interested in the power consumption of neural networks. As shown in Fig. \ref{LIF}, SNNs offer a natural solution because they have a similar way of transmitting information as the brain, using spike neurons like the leaky integrate-and-fire (LIF) neurons instead of traditional artificial neurons, such as ReLU-activated neurons. Brain-like LIF model corresponds to its physical circuit. The soma constantly receives integrated input from dendrites. When the sum of total input exceeds a certain threshold $\vartheta$, an output spike is generated and then delivered to other neurons by the axon. After firing a spike, the membrane potential is reset to $V_{rest}$. SNNs have the advantage of lower power consumption during the inference stage. Besides, SNNs have successfully permeated into a myriad of application domains such as image and speech recognition, object detection, autonomous driving and other intelligence-related real-world applications \cite{10636118, 10767300, 9138762}. Currently, there are three types of spiking learning frameworks in terms of implementation: 1) due to the discreteness of spikes and the intrinsic non-differentiability of spike firing function impeding the direct applicability of the traditional backpropagation (BP), surrogate-based BP algorithm thus can be derived by replacing the spiking function with differential functions or adding some limitation or clipping to the BP process (e.g., \cite{BP3}). 2) Based on spike-timing-dependent-plasticity (STDP) algorithm, SNN models learn each module separately (e.g., \cite{STDP2}). 3) Different from the above SNN algorithms, the ANN-to-SNN algorithms assume that SNNs have equal efficiency as the counterpart traditional neural networks such as (e.g., \cite{wu2021tandem}).

\begin{figure}[]
\centering
{\includegraphics[width = 10.5cm]{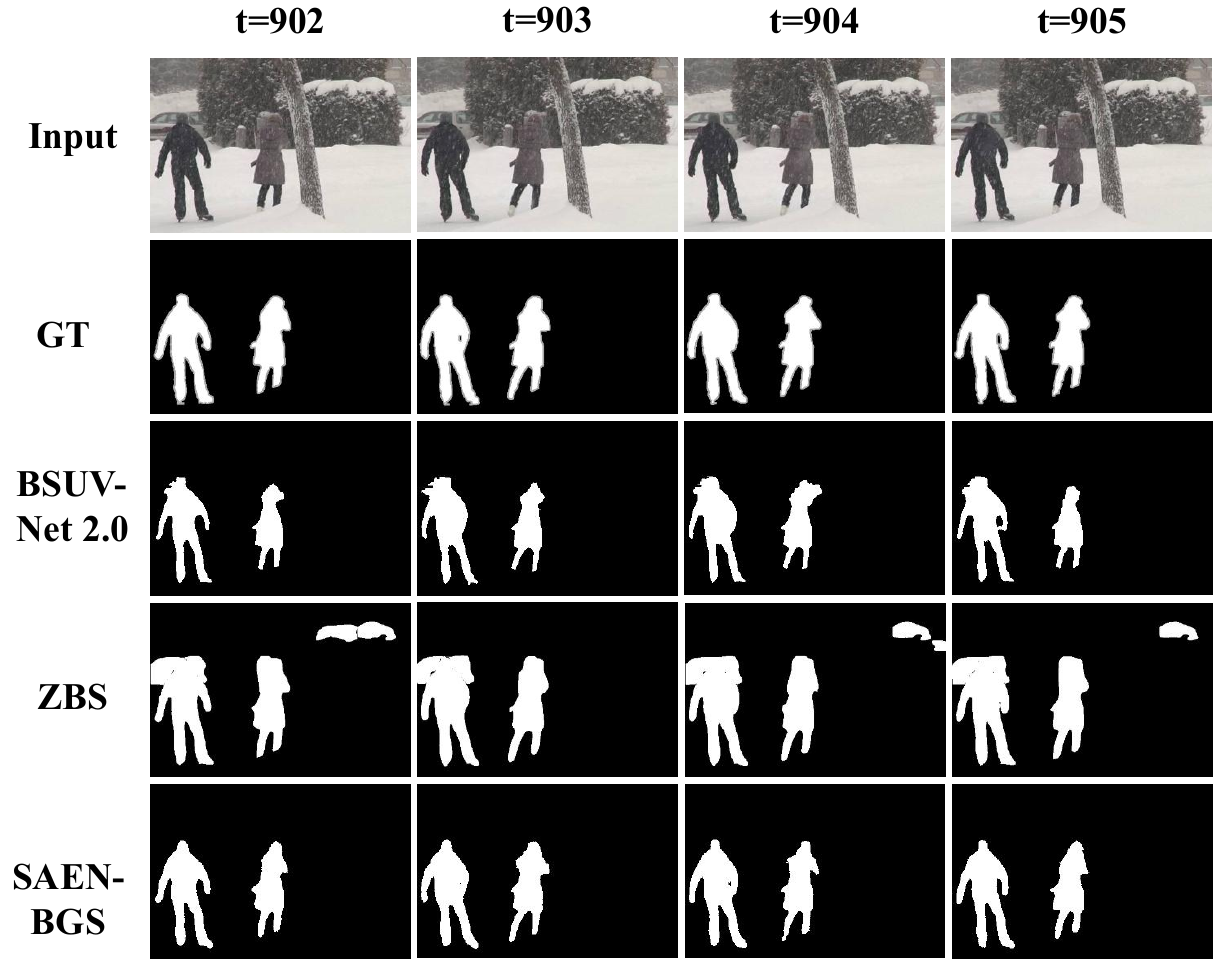}}
\caption{Separation Performance of our method and two state-of-the-art baselines over five continuous frames in the badweather video with high background noise.}
\label{temporal}
\end{figure}

\section{{Methodology}}
\subsection{Problem Formulation}
BGS serves as a fundamental research topic in video processing and has been widely investigated nowadays. Given a sequence $\boldsymbol{X}$ of consecutive frames \textit{X$_1$, \ldots, X$_N$} with a scene cluttered by various moving objects, such as cars or pedestrians, and the expected output is a sequence $\hat{\boldsymbol{X}}$ of frames \textit{$\hat{X}_1$, $\ldots$, $\hat{X}_N$} showing the backgrounds of each scene without those objects. Typically, supervised BGS methods require the assistance of the given label/mask $\boldsymbol{M}$. Despite the current BGS methods have made much progress on separation performance, it is inevitable that they still get stuck in background noises, such as lighting variation, camera moving and weather changes. Thus, they are erroneously perceived as the foremost subjects in dim lighting or shadows, regardless of alterations in the backdrop. This is verified by the experimental result in Fig. \ref{temporal}, where the continuous frames from t = 902 to t = 905 are randomly chosen. As compared with ZBS [32] and BSUV-Net 2.0 [24], snow and the car in the background are erroneously perceived as the foremost subjects, respectively. In the subsequent section, we introduce a novel SNN model aimed at enhancing separation performance. This model incorporates a continuous spiking conv-and-dconv block along with a unique self-distillation spiking learning algorithm. 

Given a SNN $\mathbf{S}$ used for BGS, its internal process is spike-event driven. First, the input frame \textit{X$_i$} needs to be encoded into spike trains $s_i$=${s_i^1, \ldots, s_i^T}$ by rate coding \cite{BP3}, where $T$ indicates the length of the encoding time window. Then the encoded spike trains $s_i$ pass over $\mathbf{S}$, and the outputs of intermediate layers are composed of spike trains and spike counts. Finally, the expected output mask can be described as $\hat{\boldsymbol{M}} = \mathbf{S}(\textit{X$_i$})$. Besides, $\mathbf{S}$ is inherently sensitive to temporal information, beneficial to alleviating the issues of current BGS methods.

\begin{figure*}[htbp]
\centering
\includegraphics[width=\linewidth]{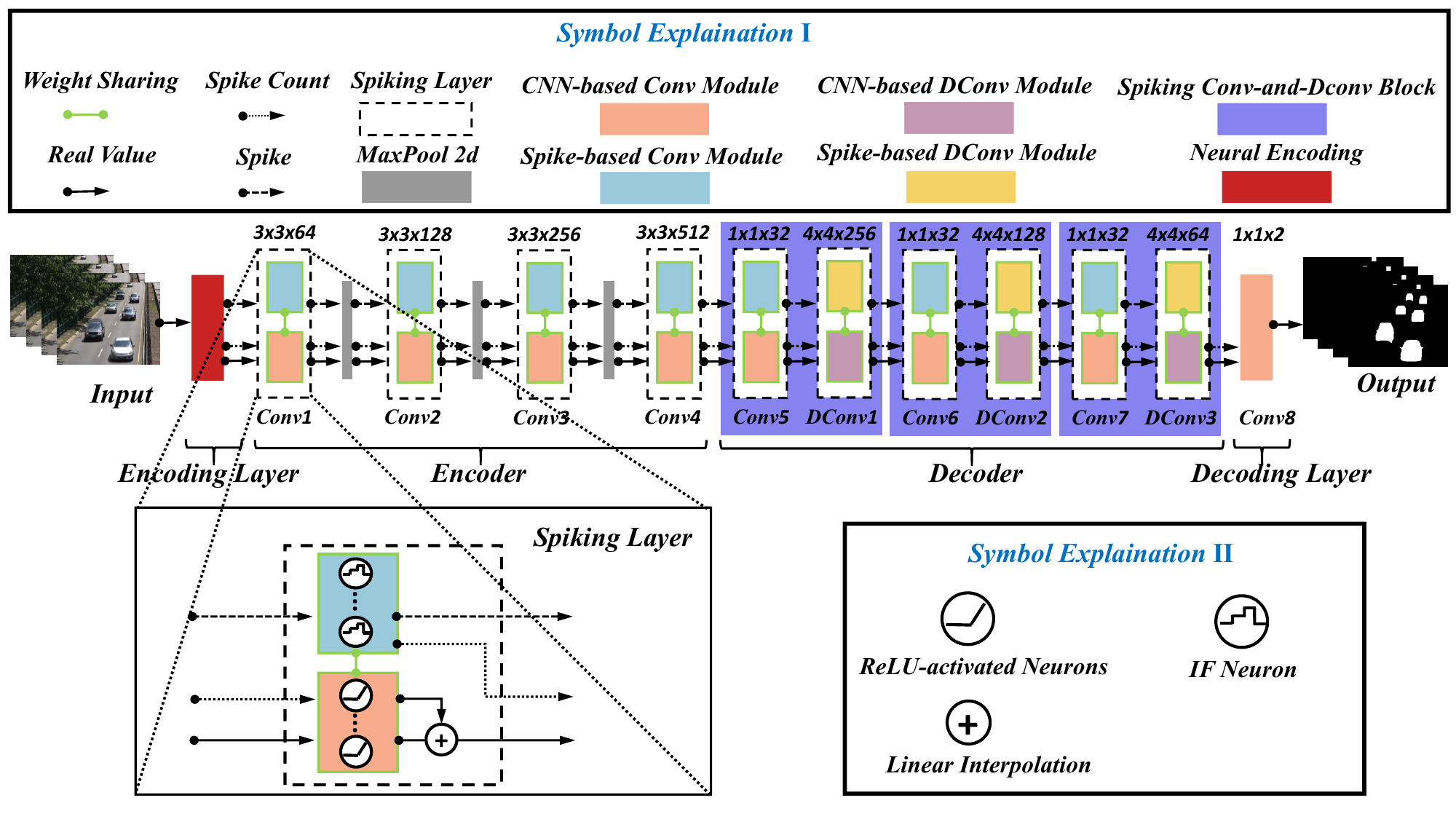}
\caption{Architecture of the proposed SAEN-BGS. Each spiking layer integrates a CNN-based module and a spike-based module with shared weights. The CNN-based module facilitates the spike-based module training but is inactive during inference. The spike-based module processes each frame in 10 time steps.}
\label{fig: Framework}
\end{figure*}

\subsection{{SAEN-BGS}}
As mentioned above, we engage to solve the existing problem that current BGS methods tend to be influenced by background noise, resulting from the insufficient extraction of temporal information by our above analysis. Additionally, the increasing number of parameters and operations in neural networks poses a potential challenge. Fortunately, SNNs have inherent sensitivity to temporal information and are friendly to low-energy computation attributed to the spike-based mechanism. Therefore, to address these issues, we apply the spike-based computation to devise a SAEN-BGS model for BGS. In detail, from the perspectives of model structure and learning strategy, we design a continuous spiking conv-and-dconv block and propose a self-distillation spiking learning algorithm, respectively. As shown in Fig. \ref{fig: Framework}, the proposed model is composed of an encoding layer, a spiking encoder, a spiking decoder and a decoding layer. CNN-based conv module and CNN-based dconv module, respectively, represent a convolutional layer and a deconvolutional layer to transmit real values rather than spikes. Details of the proposed SAEN-BGS model are described below.

\subsubsection{Neuron model}
Our model comprises two types of neurons: ReLU-activated neurons and integrate-and-fire (IF) neurons. ReLU-activated neurons, commonly used in traditional neural networks, clip activations below zero. IF neurons, inspired by biological spike-based information transmission, are utilized in SNNs, which replace LIF neurons for simplicity. Synaptic transmission between presynaptic and postsynaptic neurons is driven by the positive correlation between presynaptic spike timing and membrane potentials. In a simulation time window $N_t$, the input spikes to neuron $j$ in layer $l$ are integrated into subthreshold membrane potential $U_j^l$ at each time step $t$. Thus, the membrane potential of neuron $j$ in layer $l$ is modeled as  \cite{wu2021tandem}:
\begin{align}
    U_j^l[t] = U_j^l[t-1] + RI_j^l[t] - \vartheta S_j^l[t-1],
    \label{eq5}
\end{align}
in which $I_j^l[t] = \sum_iw_{ji}^{l-1}S_i^{l-1}[t]+b_j^l$ and $S_j^l[t] = \delta(U_j^l[t]-\vartheta)$ with the indicator function 
\begin{align}
    \delta(x) = \begin{cases}1, ~\text{if}~ x\ge 0\\ 0, ~\text{otherwise.} \end{cases}
\end{align} 
The synaptic weight $w_{ji}^{l-1}$ affects connection from presynaptic neuron $i$ in layer $l-1$ and $b_j^l$ is a constant injecting current. Moreover, $S_j^l[t-1]$ signifies the occurrence of a spike emitted by the neuron $j$ at the previous time step $t-1$, and $I_j^l[t] $ represents the synaptic current generated by the integration of incoming spike trains $t$. At time $t$, the membrane potential $U_j^l(t)$ exceeds the predefined threshold $\vartheta$ (commonly set to $\vartheta = 1$), leading to the initiation of an action potential, also known as a spike. That is:
\begin{align}
    U_j^l(t) \ge \vartheta ~, ~ \frac{dU_j^l(t)}{dt} >0.
\end{align}
 Following the generation of a spike, the membrane potential $U_j^l(t)$ is reset to the resting potential $U_{rest}$ and remains in a refractory state for a specific duration. Synaptic weight, or synapse conductance, changes according to presynaptic and postsynaptic neuronal activities. This variability, driven by synaptic plasticity, enables neuronal learning. Encoding methods are categorized into spike-rate-based (counting spikes over time) and spike-time-based (focusing on precise spike timing). Our encoding method is detailed in the next subsection.

\subsubsection{Encoding and decoding layers}
The encoding layer and the decoding layer serve as the input and output modules, respectively, in our proposed model. Time-varying input currents, treated as real-valued inputs, are directly incorporated into Eq. \ref{eq5} at each time step. This neural encoding method eliminates the sampling errors associated with rate coding, enabling precise and efficient inference \cite{BP3}. It can be denoted as $s$ = $\phi$($x$), where $s$ and $x$ indicate the produced spike trains and the real-value input, respectively. As illustrated in Fig. \ref{fig: Framework}, the spiking layer processes both spike trains and spike counts as inputs.

To facilitate pattern classification, the output spike trains from the SNNs must be decoded into distinct pattern classes. This decoding can be accomplished at the SNNs' output layer using either discrete spike counts or continuous aggregated membrane potentials (without spiking). Utilizing aggregated membrane potentials provides a smoother learning process, as it allows for the computation of continuous error gradients at the output layer \cite{wu2021tandem}. Therefore, a convolutional layer as the decoding layer is employed.

\begin{figure}[]
\centering
\includegraphics[width=12cm]{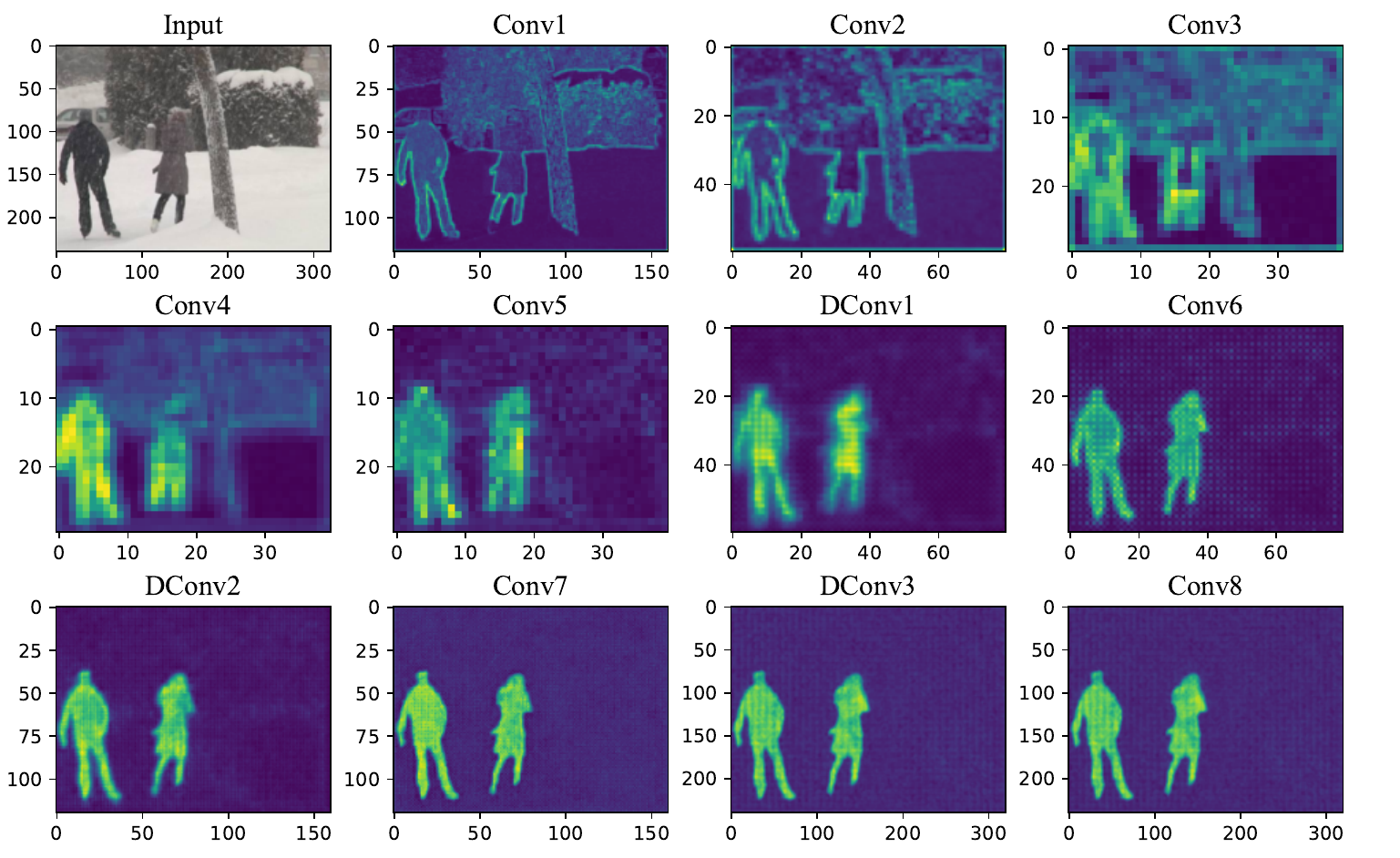} 
\caption{Outputs of intermediate layers of SAEN-BGS model over a randomly selected frame.}
\label{intermediate}
\end{figure}

\subsubsection{Spiking conv-and-dconv block}
The spiking conv-and-dconv block is designed for information compression and deconvolution for feature reconstruction, which has proven effective in various generative tasks \cite{GANs}. Specifically, we propose that applying deconvolution (dconv) after convolution (conv) filtering can help reduce background noise. The spiking conv-and-dconv block primarily consists of a spiking conv layer and a spiking dconv layer. In SAEN-BGS, the decoder is constructed using three such blocks. Each spiking layer integrates a CNN-based module for parameter training and a spike-based module for inference (testing or validation), with shared parameters, e.g., convolution kernels. In this block, the kernel size of the spiking conv layer is set to 1$\times$1$\times$$C_{out}$, where $C_{out}$ denotes the number of output channels. Since the output channel number from the preceding spiking layer exceeds that of the current spiking conv layer, the block effectively compresses channel information, which may aid in suppressing background noise. As illustrated in Fig. \ref{intermediate}, the results indicate that the background noise in the outputs of the proposed conv-and-dconv spiking blocks is significantly reduced compared to that in the front encoder, thus confirming the effectiveness of our designed block.

\begin{figure}[]
\centering
\includegraphics[width=11cm]{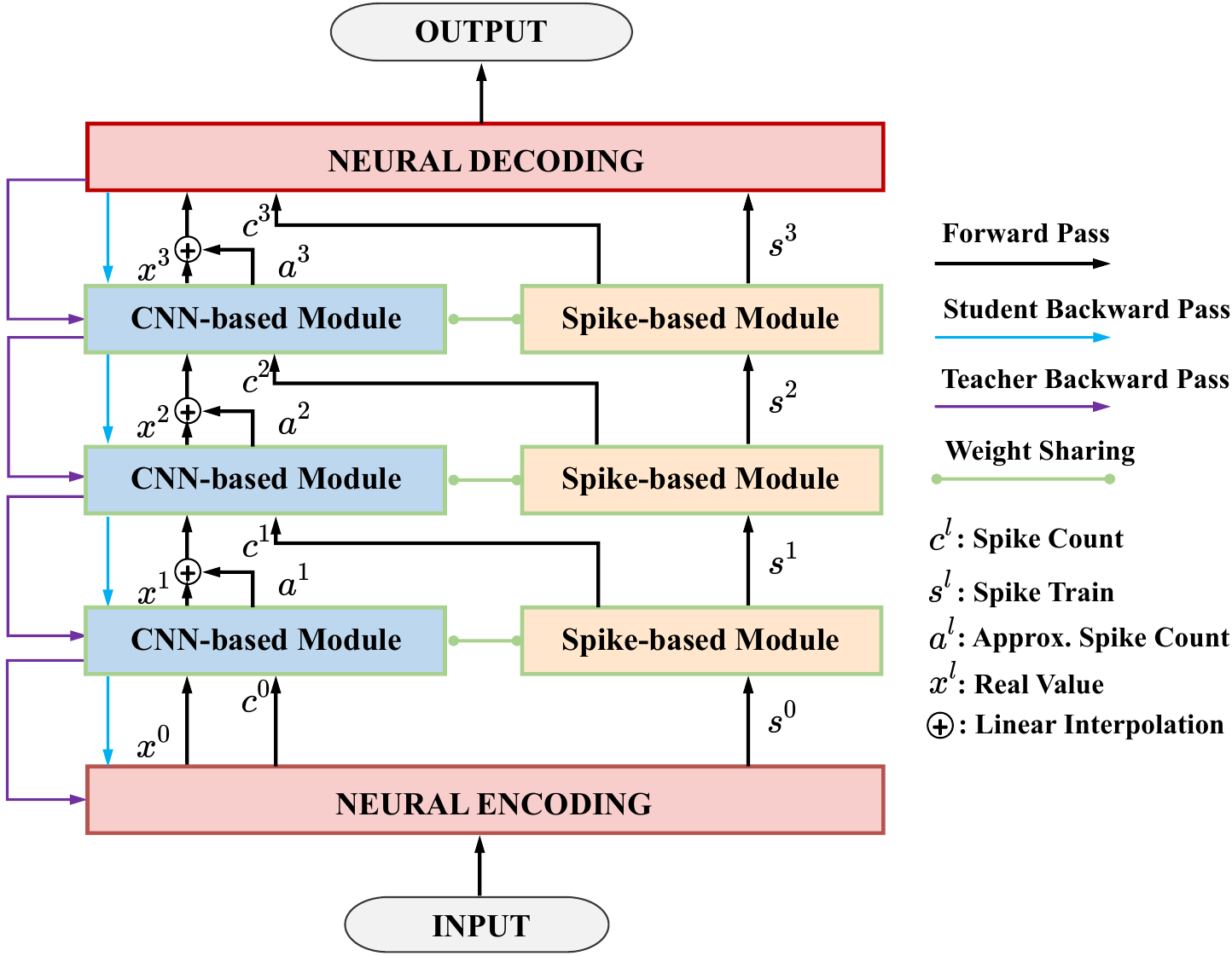} 
\caption{{Frameworks of our proposed self-distillation spiking learning algorithm.}}
\label{fig:Joint-framework}
\end{figure}

\subsection{Self-Distillation Spiking Learning Algorithm}
\begin{figure}[]
\centering
\includegraphics[width=12cm]{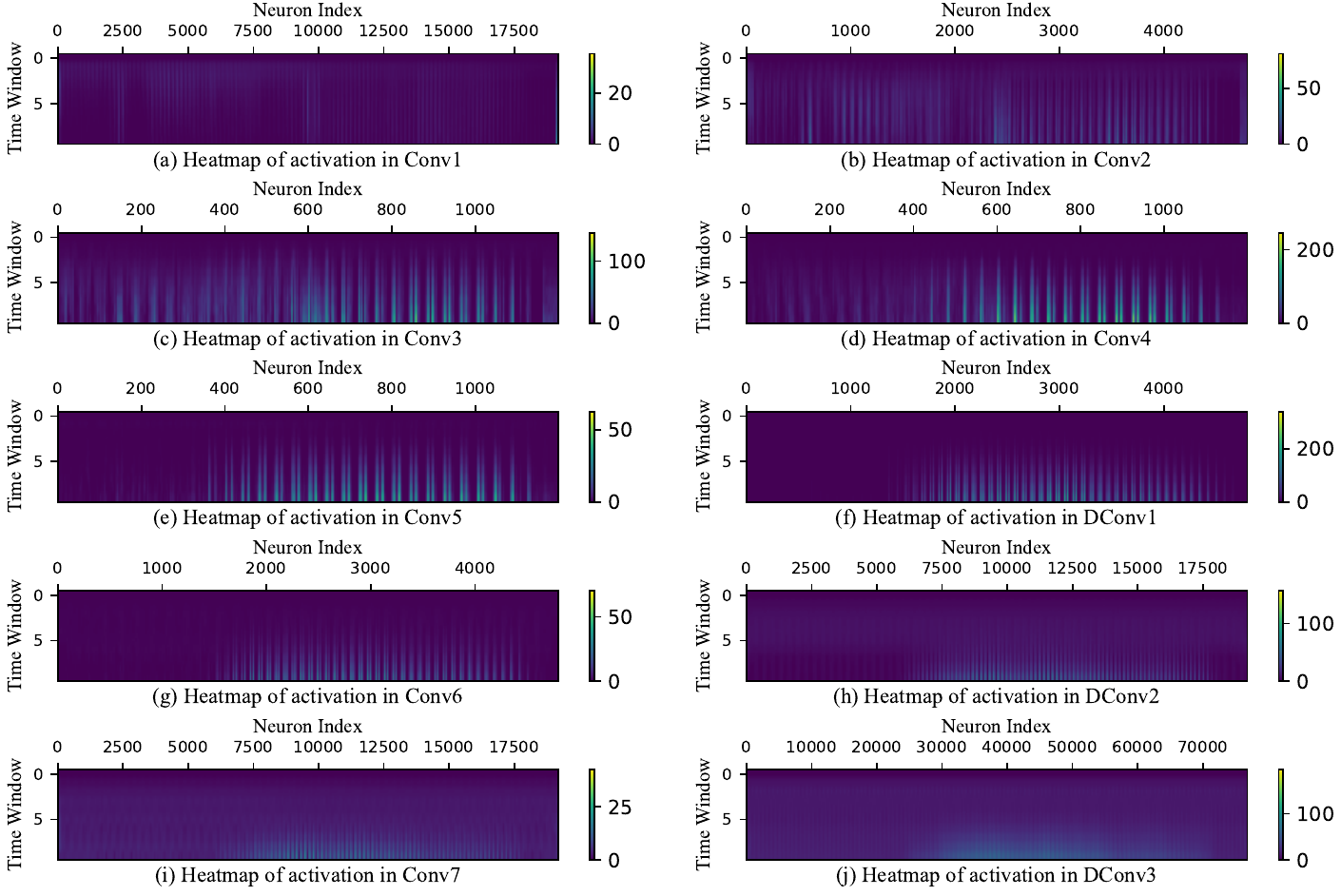} 
\caption{Illustration of activation heatmap of spiking layer in our SAEN-BGS over an input.}
\label{heatmap}
\end{figure}

Our proposed algorithm aims to provide energy-efficient learning for our SAEN-BGS, inspired by the ANN-to-SNN framework in \cite{wu2021tandem}, where the CNN-based module serves as the teacher overseeing the SNN-based counterpart in its learning process, assuming that the spike counts from the spike-based module can be accurately estimated by the actual values from the CNN-based module. However, approximation loss exists along with performance degradation, especially in noisy circumstances. In addition, it is not advisable to make sweeping generalizations to new data sourced from the identical distribution. To address that, as shown in Fig. \ref{fig:Joint-framework}, a learning pathway is built to independently pass over every CNN-based module via linear interpolation $\eta x^l+(1-\eta) a^l$ of real-valued data $x^l$ and approximated spike count $a^l$, where the hyperparameter $\eta$ is set 0.5 as default. This is motivated by the multi-task learning \cite{multi-task_ML} and the self-distillation learning mechanism \cite{self-distill}. Concretely, this design aims to further boost the robustness of spike-based modules to noise by using the approximated spike count from the CNN-based module, and reduce the bias of the CNN-based module in the learning process by the new learning pathway over every CNN-based module. Since it can be seen as increasing new auxiliary sub-networks within the network itself, the learning process can be termed as self-distillation in multi-task views, so-called self-distillation spiking learning algorithm. The main task is to train the spike-based module by sharing weights from the CNN-based module, while the related task is to train the CNN-based module itself. As ahown in Fig. \ref{heatmap}, we illustrate the heatmap of activation in all spiking layers, which presents the sparsity of activation on spike neurons directly. More experimental results are presented at Sec. \ref{experiment}.

{For the BGS problem, our goal is to minimize the negative log-likelihood loss between the ground-truth mask $\boldsymbol{M}$ and the estimated outputs $\hat{\boldsymbol{M}}$. Given frames \textit{X$_1$, \ldots, X$_N$} with labels \textit{M$_1$, \ldots, M$_N$}, the SAEN-BGS outputs are denoted as \textit{$\hat{M}_1$, \ldots, $\hat{M}_N$}. The loss function $\mathcal{L}$ can be formulated as:
\begin{align}
    \mathcal{L}(\hat{\boldsymbol{M}}, \boldsymbol{M}) =-\frac{1}{N}\sum\limits_{i=1}^{N} M_i\log\hat{M}_i. \label{loss_mse}
\end{align}
}

As mentioned above, to achieve the multi-task learning, the overall loss function $\mathcal{L}_{all}$ is applied, where $\hat{\boldsymbol{M}}$ and $\hat{\boldsymbol{M}}^r$, represent the outputs of main task (along the pathway of $s^i$ and $c^i$) and related task (along the pathway of $x^i$), respectively. The hyperparameter $\alpha$ is equal to 0.8:
\begin{align}
    \mathcal{L}_{main}=\mathcal{L}(\hat{\boldsymbol{M}}, \boldsymbol{M}),
    \label{loss_main}
\end{align}
\begin{align}
    \mathcal{L}_{aux}=\mathcal{L}(\hat{\boldsymbol{M}}^r, \boldsymbol{M}),
    \label{loss_related}
\end{align}
\begin{align}
    \mathcal{L}_{all}=\alpha\mathcal{L}_{main}+(1-\alpha)\mathcal{L}_{aux}.
    \label{joint-loss-mse}
\end{align}

As for the non-differentiable nature of spike generation, the error backpropagation (BP) method cannot be directly applicable to the training process. To that end, we apply the ANN-to-SNN rule by spike count approximation. Next, the spike count from neuron $i$ at layer $l$ is formulated as:
\begin{align}
    c_i^l = \sum\limits_{t=1}^{N_t} S_i^l[t].
\end{align}

The relationship between ANN activation functions and SNN spike counts is established through shared weights, modeled mathematically as:
\begin{align}
    \Delta t = \rho(\frac{\vartheta}{U_j^l/N_t}),
\end{align}
in which the aggregated action potential of neuron $j$ in layer $l$ is written as $ U_j^{l} = \sum_i w_{ji}^{l-1}c_i^{l-1} +b_j^lN_t$. $\vartheta$, $\rho$ and $\Delta t$ denote the firing threshold, the non-linear transformation of ReLU-activated neurons, and the time step, respectively. Hence, the approximated spike count $a_j^l$ is obtained as:
\begin{align}
    a_j^l = \frac{N_t}{\Delta t} = \frac{1}{\vartheta}\rho(\sum_iw_{ji}^{l-1}c_i^{l-1} + b_j^lN_t).
    \label{a_j^l}
\end{align}

\begin{algorithm}
\begin{algorithmic}
\REQUIRE{Training frames $\boldsymbol{X}$; Ground truth masks $\boldsymbol{M}$; Total epoch $\mathbf{E}$; SAEN-BGS $\mathbf{F}$ = \{Neural Encoding $\phi$, Encoder $\mathbf{F}_{en}$ = (Conv1, Conv2, Conv3, Conv4), Decoder $\mathbf{F}_{de}$ = ($\langle$Conv5, DConv1$\rangle$, $\langle$Conv6, DConv2$\rangle$, $\langle$Conv7, DConv3$\rangle$), Decoding Layer Conv8\} with parameter $\theta_\mathbf{F}$, $\langle \cdot \rangle$ denotes Spiking Conv-and-DConv Block; Learning rate $\gamma$; Hyperparameter $\alpha$.}      
\ENSURE {Model parameters $\theta_\mathbf{F}^*$.}       
\BlankLine
\For{$e\leftarrow 1$ \KwTo $\mathbf{E}$}{
\For{$i\leftarrow 1$ \KwTo $len(\boldsymbol{X})$}{
\tcp{Extract training data }            
$x \leftarrow \boldsymbol{X}[i]$,
$y \leftarrow \boldsymbol{M}[i]$\;\\
\tcp{In encoding layer}
$s^0$, $c^0$, $x^0$, $a^0 \leftarrow \phi$($x$), $x$, $x$, 0 (as in Eq. \ref{eq5})\;\\
\tcp{In spiking layers}
\For{$\ell \leftarrow 1$ \KwTo $10$}{
${x'}^{l-1}\leftarrow \eta x^{l-1}+(1-\eta)a^{l-1} $ \;\\
$s^{\ell}$, $c^{\ell}$, $x^{\ell}$, $a^{\ell} \leftarrow \mathbf{F}[\ell]$($s^{\ell-1}$, $c^{\ell-1}$, ${x'}^{\ell-1}$) \;\\
}
\tcp{In decoding layer}
$\hat{y}, \hat{y}^r \leftarrow \text{Conv8}$($c^{10}, \eta x^{10}+(1-\eta)a^{10}$) \;\\
\tcp{Construct loss functions}
$\mathcal{L}_{main}\leftarrow Eq. \ref{loss_main}$ with input of $\hat{y}$ and $y$ \;\\
$\mathcal{L}_{aux}\leftarrow Eq. \ref{loss_related}$ with input of  $\hat{y}^r$ and $y$ \;\\
$\mathcal{L}_{all}\leftarrow \alpha\mathcal{L}_{main}+(1-\alpha)\mathcal{L}_{aux}$ (as in Eq. \ref{joint-loss-mse}) \;\\
\tcp{Update parameters of SAEN-BGS}
$\theta_\mathbf{F}$$\leftarrow$$\theta_\mathbf{F}$-$\gamma$$\nabla_{\theta_\mathbf{F}}{\mathcal{L}_{all}}$  \;\\
$\theta_\mathbf{F}^*$$\leftarrow$$\theta_\mathbf{F}$
}
}
\end{algorithmic}
\caption{Self-Distillation Spiking Learning}
\label{alg:alg1}
\end{algorithm}

Thus, $a_j^l$ is determined by the ReLU activation in the CNN-based module, using the spike count $c_i^{l-1}$ as input and $b_j^lN_t$ as the bias term. This simplification enables error gradient estimation at the spike-train level using CNN-based modules, following the tandem learning rule with firing rates instead of spike timings \cite{wu2021tandem}.

The overall process of self-distillation spiking learning is shown in Algorithm \ref{alg:alg1}. In a training epoch, training frames $\boldsymbol{X}$ as input to SAEN-BGS $\mathbf{F}$ is first encoded into spike trains $s^0$ and spike counts $c^0$ at encoder layer of $\mathbf{F}$ obeyed by Eq. \ref{eq5}. It notes that the original $\boldsymbol{X}$ is also copied as one of the outputs of the encoding layer, namely $x^0$, to pass the CNN-based module of the first spiking layer. Then the resulting ten spiking layers all receive the spike trains $s^{\ell-1}$, spike counts $c^{\ell-1}$, real-value output $x^{\ell-1}$ and approximated spike count $a^{\ell-1}$ from the previous blocks and output the new spike trains $s^{\ell}$, spike counts $c^{\ell}$, real-value output $x^{\ell}$ and approximated spike count $a^{\ell}$. Besides, the decoding layer processes the spike counts and real-value output from the last spiking layer in the decoder and outputs the mask predictions $\hat{\boldsymbol{M}}$, $\hat{\boldsymbol{M}}^r$. Finally, we construct the loss function using Eq. \ref{joint-loss-mse} and update the parameters of SAEN-BGS by SGD.

\section{Experiments}
\label{experiment}
\subsection{Experimental Setup}
We present the experimental setup for the CDnet-2014~\cite{6910011} and DAVIS-2016~\cite{davis2016-dataset} datasets. The proposed neural networks are implemented using the PyTorch framework~\cite{pytorch}, with the negative log-likelihood loss as the objective function and RMSprop as the optimizer. It should be noted that in addition to our SAEN-BGS, we also display the performance of the counterpart CNNs (AEN-BGS) to demonstrate the energy efficiency of our proposed self-distillation spiking learning constantly.

\subsubsection{Baselines}
To verify the segmentation performance and efficient energy consumption of our proposed SAEN-BGS, we compare it with the counterpart AEN-BGS, and several state-of-the-art methods, including  ZBS\cite{zbs}, CwisarDH \cite{6910014}, Spectral-360 \cite{6910013}, FTSG  \cite{6910016}, SuBSENSE \cite{6910015}, DeepBS \cite{DeepBS}, SemanticBGS \cite{SemanticBGS}, IUTIS-5 \cite{IUTIS}, BSUV-Net \cite{Tezcan2020}, IUTIS-5+SemanticBGS \cite{Tezcan2020}, BSUV-Net 2.0 \cite{Tezcan2021}, WisenetMD \cite{WisenetMD}, RTSS \cite{RTSS}, SWCD \cite{SWCD} and PAWCS \cite{PAWCS}.

\subsubsection{Training details} We evaluate our SAEN-BGS and other baselines on CDnet-2014 and DAVIS-2016 datasets. The training details for our SAEN-BGS and its counterpart AEN-BGS, are as follows. It should be noted that except for AEN-BGS, other baselines keep their original setting.

The CDnet-2014 dataset comprises 53 videos across 11 categories, all used to evaluate our proposed SAEN-BGS. In consistency with \cite{Tezcan2021}, we also follow the two training schemes employed: a small training set with 200 randomly selected frames from each video and a large training set with randomly selected 70\% of the frames from each video. Additionally, since the video frame dimensions vary from 240$\times$240 to 526$\times$720, all frames and ground truth are resized to 240$\times$320 for simplicity. The batch size is set to 8, with training over 100 epochs. As for the learning rate (LR), our SAEN-BGS sets 0.0005 to the night videos and camera jitter categories except for other categories using an initial LR of 0.0002. For its counterpart AEN-BGS,  LR is set to 0.001, except for the night videos and PTZ categories under the large training set with an initial LR of 0.0002. Regarding LR adapter techniques, our SAEN-BGS uses the MultiStepLR scheduler for the intermittent object motion category and the ReduceLROnPlateau scheduler for night videos in the small training set and the PTZ category, while the StepLR scheduler is used for all other videos. Then the counterpart AEN-BGS utilizes the ReduceLROnPlateau scheduler as an early stopping strategy, except for night videos in the large training set and the PTZ category in the small training set applying the StepLR scheduler. Additionally, our SAEN-BGS only spends a time window of 10 ms with a time step of 1 ms.

Over the DAVIS-2016 dataset, there exist almost no BGS methods evaluating. Thus, we train on the large training set scheme with randomly selected 70\% of the frames from each video, aimed to validate the overall performance. All frames and ground truths are resized to 240×320. The batch size is set to 8, over 200 training epochs and LR is set to 0.001 for AEN-BGS and 0.0002 for SAEN-BGS. For LR adapter, AEN-BGS uses the StepLRMulti scheduler, while SAEN-BGS employs the StepLR scheduler. The time window length (10 ms) and time step (1 ms) remain consistent with those used for the CDnet-2014 dataset.

\begin{figure*}[htb]
\centering
\includegraphics[width=12cm]{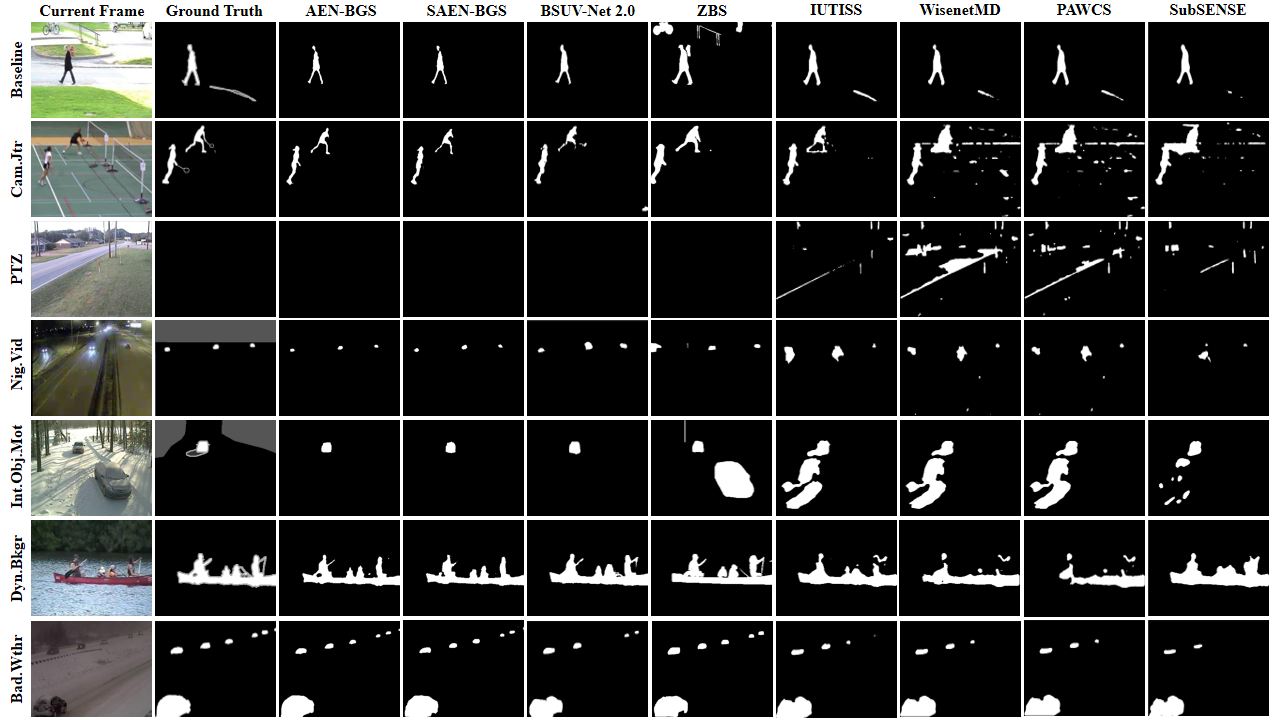}
\caption{Visual comparison of top BGS algorithms on sample frames from different categories of CDNet-2014.}
\label{fig: CD2014}
\end{figure*}

\subsubsection{Basic evaluation metrics}
To assess the performance of BGS, seven evaluation metrics are employed: False Positive Rate (FPR), False Negative Rate (FNR), Recall (Rec), Precision (Pre), Specificity (Spe), Percentage of Wrong Classifications (PWC), and F-measure (Fm). These metrics are derived from the definitions of true positive (TP), false positive (FP), false negative (FN), and true negative (TN). Specifically, TP and FP represent pixels correctly and incorrectly classified as foreground, respectively, while TN and FN denote pixels correctly and incorrectly classified as background. The seven metrics are defined as follows:
\begin{align}
    &\text{FPR} = \frac{FP}{FP+TN},\text{ }
    \text{FNR} = \frac{FN}{TP+FN},\notag\\
    &\text{Rec} =\frac{TP}{TP+FN}, \text{ }
    \text{Pre} = \frac{TP}{TP+FP},\notag\\
    &\text{Spe} = \frac{TN}{TN+FP}, \text{ }
     \text{PWC} = \frac{(FN+FP)}{TP+FN+FP+TN}, \notag\\
    &\text{Fm} = \frac{2\times \text{Pre}\times \text{Rec}}{\text{Pre} + \text{Rec}}.
\end{align}
Clearly, lower FPR, FNR, and PWC values indicate better performance, while higher Rec, Pre, Spe, and Fm reflect superior results. Except for PWC, the units of all other metrics are \%.

\begin{table*}[htb]
\centering
\caption{Evaluation results on CDnet 2014 dataset under the small training set (randomly selected 200 frames per video) among different compared models using several evaluation metrics. \textbf{\textcolor{blue}{Blue}} indicates the best. \textbf{\textcolor{red}{Red}} indicates the second.}
\resizebox{\linewidth}{!}{
\begin{tabular}{c|c|c|c|c|c|c|c}
\hline
\textbf{Model} & \textbf{Rec ($\uparrow$)} & \textbf{Spe ($\uparrow$)} & \textbf{FPR ($\downarrow$)}    & \textbf{FNR ($\downarrow$)}    & \textbf{PWC ($\downarrow$)}    & \textbf{Fm ($\uparrow$)} & \textbf{Pre ($\uparrow$)} \\ \hline
\textbf{AEN-BGS}  & {\textbf{\textcolor{red}{91.97}}} & \textbf{\textcolor{blue}{99.84}}      & \textbf{\textcolor{blue}{0.12}} & {\textbf{\textcolor{red}{8.01}}} & \textbf{\textcolor{blue}{0.296}} & \textbf{\textcolor{blue}{92.34}}    & \textbf{\textcolor{blue}{92.74}}    \\ \hline
\textbf{SAEN-BGS}  & {89.59} & {\textbf{\textcolor{red}{99.78}}}      & {\textbf{\textcolor{red}{0.20}}} & {10.38} & {\textbf{\textcolor{red}{0.388}}} & {\textbf{\textcolor{red}{90.12}}}    & \textbf{\textcolor{red}{90.71}}    \\ \hline
CwisarDH \cite{6910014}       & 66.08          & 99.48               & 0.52          & 33.92          & 1.527          & 68.12             & 77.25             \\ \hline
Spectral-360 \cite{6910013}   & 73.45          & 98.61               & 1.39          & 26.55          & 2.272          & 67.32             & 70.54             \\ \hline
FTSG  \cite{6910016}         & 76.57          & 99.22               & 0.78          & 23.43          & 1.376          & 72.83             & 76.96             \\ \hline
SuBSENSE \cite{6910015}      & 80.70          & 98.84               & 1.16          & 19.30          & 1.842          & 73.31             & 74.63             \\ \hline

CascadeCNN \cite{CascadeCNN}     &\textbf{\textcolor{blue}{95.06}}          & 99.68               & 0.32          & \textbf{\textcolor{blue}{4.94}}          & 0.405          & 90.09             & 89.97             \\ \hline
DeepBS \cite{DeepBS}         & 75.45          & 99.05               & 0.95          & 24.55          & 1.992          & 74.58             & 83.32             \\ \hline
SemanticBGS \cite{SemanticBGS}   & 78.90          & 99.61               & 0.39          & 21.10          & 1.072          & 83.05             & 78.92             \\ \hline
IUTIS-5 \cite{IUTIS}       & 78.49          & 99.48               & 0.52          & 21.51          & 1.199          & 80.87             & 77.17             \\ \hline
\end{tabular}
}
\label{table: CD2014_small}
\end{table*}

\begin{table*}[htb]
\centering
\caption{ Evaluation results on CDnet 2014 dataset under the large training set (randomly selected 70\% frames per one video) among different compared models using several evaluation metrics. \textbf{\textcolor{blue}{Blue}} indicates the best. \textbf{\textcolor{red}{Red}} indicates the second.}
\resizebox{\linewidth}{!}{
\begin{tabular}{c|c|c|c|c|c|c|c}
\hline
\textbf{Model} & \textbf{Rec ($\uparrow$)} & \textbf{Spe ($\uparrow$)} & \textbf{FPR ($\downarrow$)}    & \textbf{FNR ($\downarrow$)}    & \textbf{PWC ($\downarrow$)}    & \textbf{Fm ($\uparrow$)} & \textbf{Pre ($\uparrow$)} \\ \hline
\textbf{AEN-BGS}  & \textbf{\textcolor{blue}{94.69}} & \textbf{\textcolor{blue}{99.91}}      & \textbf{\textcolor{blue}{0.07}} & \textbf{\textcolor{blue}{5.29}} & \textbf{\textcolor{blue}{0.162}} & \textbf{\textcolor{blue}{94.60}}    & \textbf{\textcolor{blue}{94.56}}    \\ \hline
\textbf{SAEN-BGS}  & {\textbf{\textcolor{red}{92.92}}} & {\textbf{\textcolor{red}{99.88}}}      & {\textbf{\textcolor{red}{0.11}}} & {\textbf{\textcolor{red}{7.05}}} & {\textbf{\textcolor{red}{0.229}}} & {\textbf{\textcolor{red}{92.82}}}    & {\textbf{\textcolor{red}{92.74}}}    \\ \hline
ZBS \cite{zbs}  & {84.03} & {99.81}      & {0.19} & {15.97} & {0.563} & {85.15}    & {88.02}    \\ \hline
IUTIS-5 \cite{IUTIS}       & 78.49          & 99.48               & 0.52          & 21.51          & 1.199          & 80.87             & 77.17             \\ \hline
BSUV-Net \cite{Tezcan2020}       & 82.03          & 99.46               & 0.54          & 17.97          & 1.140          & 78.68             & 81.13             \\ \hline
IUTIS-5+SemanticBGS \cite{Tezcan2020}      & 78.90          & 99.61               & 0.39          & 21.10          & 1.072          & 78.92             & 83.05             \\ \hline
BSUV-Net+SemanticBGS  \cite{Tezcan2020}         & 81.79          & 99.44               & 0.56          & 18.21          & 1.133          & 79.86             & 83.19             \\ \hline
BSUV-Net 2.0 \cite{Tezcan2021}       & 81.36          & 99.79               & 0.21          & 18.64          & 0.761          & 83.87             & 90.11             \\ \hline
Fast BSUV-Net 2.0 \cite{Tezcan2021}   & 81.81          & 99.56               & 0.44          & 18.19          & 0.905          & 80.39             & 84.25             \\ \hline
WisenetMD \cite{WisenetMD}       & 81.79          & 99.04               & 0.96          & 18.21          & 1.614          & 76.68             & 75.35             \\ \hline
\end{tabular}
}
\label{table: CD2014_large}
\end{table*}

\subsubsection{Energy evaluation metrics}
To verify the energy efficiency of our SAEN-BGS learned by our proposed self-distillation spiking learning algorithm, we compute the energy used in model inference according to~\cite{kim2020revisiting}, which is a common way to compare the energy consumption in SNNs. The computational cost is measured by the total number of Floating-Point Operations (FLOPs), which corresponds to matrix-vector multiplications and scales proportionally. Note that energy calculations focus on the SNN model during the test phase, where CNN-based modules are inactive. For each layer $l_i$, the FLOPs of the CNN-based module are given by:
\begin{align}
\begin{split}
    &FLOPs_{ANN\_Module}(l_i) = \\
    &\begin{cases}
    k^2 \times O^2  \times C_{in}  \times C_{out}, ~\text{if $l_i$ denotes the convolutional layer,}~
    \\ C_{in}  \times C_{out}, ~\text{if $l_i$ denotes the linear layer}, \end{cases} 
    \label{flops_ann}
\end{split}
\end{align}
where $k$ and $O$ denote the kernel size and output feature map size, respectively, while $C_{in}$ and $C_{out}$ represent the channel numbers of the input and output. To compute the FLOPs of the spike-based module for each spiking layer $l_i$, the spiking rate $R_s(l_i)$, as SNN consumes energy only during spike firing, can be defined as:
\begin{align}
     R_s(l_i) = \frac{\#\text{spikes per layer $l_i$ over all time steps}}{\#\text{neurons per layer $l_i$}},
    \label{spiking_rate}
\end{align}
i.e., the average firing rate per neuron. For the whole SNN, the average spike rate (AvR$_s$) is equal to $\frac{1}{11}\sum\limits_{l=1}^{11} {R}_s(l)$. Therefore, FLOPs for the spike-based module per spiking layer are calculated as:
\begin{align}
\begin{split}
    FLOPs_{Spike\_Module}(l_i) = FLOPs_{ANN\_Module}(l_i) \times R_s(l_i).
    \label{flops_snn}
\end{split}
\end{align}
Thus, the total inference energy consumption for the CNN-based module ($E_{ANN\_Module}$) and the spike-based module ($E_{Spike\_Module}$) is calculated as:
\begin{align}
         E_{ANN\_Module} = \sum\limits_{l_i} FLOPs_{ANN\_Module}(l_i) \times E_{MAC},
    \label{E_ANN}
    \end{align}
    \begin{align}
         E_{Spike\_Module} = \sum\limits_{l_i} FLOPs_{Spike\_Module}(l_i) \times E_{AC},
    \label{E_SNN}
    \end{align} 
where $E_{AC}$, $E_{MAC}$ are derived from a standard 45nm Complementary Metal–Oxide–Semiconductor (CMOS) process, with
$E_{MAC}$ = 4.6$pJ$ and $E_{AC}$ = 0.9$pJ$ for 32 bit FP \cite{cms2014}. Here, we also utilize AvR$_s$ as the other metric to represent power consumption.

\begin{table*}[htb]
\centering
\caption{F-measure comparison of different BGS algorithms according to the per-category on CDNet-2014 dataset. \textbf{\textcolor{blue}{Blue}} indicates the best. \textbf{\textcolor{red}{Red}} indicates the second.}
\resizebox{\linewidth}{!}{
\begin{tabular}{c|c|c|c|c|c|c|c|c|c|c|c|c}
\hline
\textbf{Model} &
\textbf{\makecell[c]{Bad.\\Wthr}} & 
\textbf{\makecell[c]{Low.\\Frm.\\Rate}} & 
\textbf{\makecell[c]{Nig.\\Vid}}    & 
\textbf{PTZ}    & 
\textbf{Thrml}    & 
\textbf{Shadow} &
\textbf{\makecell[c]{Int.\\Obj.\\Mot}} &
\textbf{\makecell[c]{Cam.\\Jtr}} &
\textbf{\makecell[c]{Dyn.\\Bkgr}} &
\textbf{\makecell[c]{Base\\-line}} &
\textbf{\makecell[c]{Turb}} &
\textbf{Overall} 
\\ \hline

\textbf{AEN-BGS}  & \textbf{\textcolor{blue}{96.44}} & \textbf{\textcolor{blue}{94.56}}      & \textbf{\textcolor{blue}{90.75}} & \textbf{\textcolor{blue}{87.76}} & \textbf{\textcolor{blue}{97.53}} & 95.83    & \textbf{\textcolor{blue}{94.54}} & \textbf{\textcolor{blue}{96.75}} & \textbf{\textcolor{blue}{96.28}} & \textbf{\textcolor{blue}{97.47}} & \textbf{\textcolor{blue}{92.75}}& \textbf{\textcolor{blue}{94.60}}\\ \hline

\textbf{SAEN-BGS}  & {\textbf{\textcolor{red}{95.67}}} & {\textbf{\textcolor{red}{93.16}}}      & {\textbf{\textcolor{red}{87.67}}} & {\textbf{\textcolor{red}{85.88}}} & {\textbf{\textcolor{red}{95.51}}} & 93.83    & \textbf{\textcolor{red}{90.60}} & \textbf{\textcolor{red}{96.22}} & \textbf{\textcolor{red}{94.96}} & 96.49 & {\textbf{\textcolor{red}{91.34}}}& {\textbf{\textcolor{red}{92.85}}}\\ \hline
ZBS \cite{zbs}       & 92.29          & 74.33               & 68.00          & 81.33          & 86.98               & \textbf{\textcolor{blue}{97.65}}            & 87.58              & 95.45    
  & 92.90           & 96.53           & 63.58
  & 85.15\\ \hline
BSUV-Net \cite{Tezcan2020}       & 87.13          & 67.97               & 69.87          & 62.82          & 85.81               & 92.33            & 74.99              & 77.43    
  & 79.67           & 96.93           & 70.51
  & 78.68\\ \hline
  
BSUV-Net+SemanticBGS  \cite{Tezcan2020}         & 87.30          & 67.88               & 68.15          & 65.62          & 84.55          & \textbf{\textcolor{red}{96.64}}             & 76.01  
& 77.88  & 81.76    & 96.40   & 76.31  & 79.86          \\\hline

BSUV-Net 2.0 \cite{Tezcan2021}       & 88.44          & 79.02              & 58.57          & 70.37          & 89.32          & 95.62             & 82.63   & 90.04    & 90.57 & 96.20 & 81.74 & 83.87     \\ \hline

Fast BSUV-Net 2.0 \cite{Tezcan2021}   & 89.09          & 78.24               & 65.51          & 50.14          & 83.79          & 88.90             & 90.16            & 88.28
  &  73.20     & \textbf{\textcolor{red}{96.94}}    & 79.98   & 80.39\\ \hline

IUTIS-5+SemanticBGS
\cite{Tezcan2020}      & 82.60          & 78.88               & 50.14          & 56.73         & 82.19          & 94.78             & 78.78      & 83.88  & 94.89 & 96.04 & 69.21 & 78.92             \\ \hline

IUTIS-5 \cite{IUTIS}       & 82.48          & 77.43               & 52.90          & 42.82          & 83.03          & 90.84             & 72.96             & 83.32
  & 89.02             & 95.67             & 78.36
   & 77.17\\ \hline
WisenetMD \cite{WisenetMD}       & 86.16          & 64.04               & 57.01          & 33.67          & 81.52          
    & 89.84             & 72.64        & 82.28
    & 83.76             & 94.87        & 83.04
    & 76.68\\ \hline

RTSS \cite{RTSS}       & 86.62          & 67.71               & 52.95          & 54.89          & 85.10        
& 95.51             & 78.64         & 83.96      
& 93.25            & 95.97         & 76.30
& 79.17\\ \hline

SWCD \cite{SWCD}       & 82.33          & 73.74                            & 58.07          & 45.45          & 85.81                & 87.79            & 70.92       & 74.11
             & 86.45          & 92.14         & 77.35
             & 75.83\\ \hline

PAWCS \cite{PAWCS}       & 81.52          & 65.88                            & 41.52          & 46.15          & 83.24                & 89.13          & 77.64          & 81.37
              & 89.38          & 93.97          & 64.50
              & 74.03\\ \hline

\end{tabular}
}
\label{table: CD2014_per_category}
\end{table*}

\begin{table}[htb]
\centering
\caption{Results comparison of different approaches on DAVIS-2016 dataset. \textbf{\textcolor{blue}{Blue}} indicates the best. \textbf{\textcolor{red}{Red}} indicates the second.}
\resizebox{0.28\linewidth}{!}{
\begin{tabular}{c|c}
\hline
{\textbf{Model}}                         & \textbf{Fm ($\uparrow$)}             \\ \hline
\textbf{AEN-BGS}                & \textbf{\textcolor{blue}{87.40}}        \\ \hline
\textbf{SAEN-BGS}               & \textbf{\textcolor{red}{85.20}}                 \\ \hline
ZBS \cite{zbs}          & 69.04                 \\ \hline
BSUV-Net 2.0 \cite{Tezcan2021}          & 63.62
\\ \hline
\end{tabular}}
\label{DAVIS-2016}
\end{table}

\begin{figure}[htb]
\centering
\includegraphics[width=9cm]{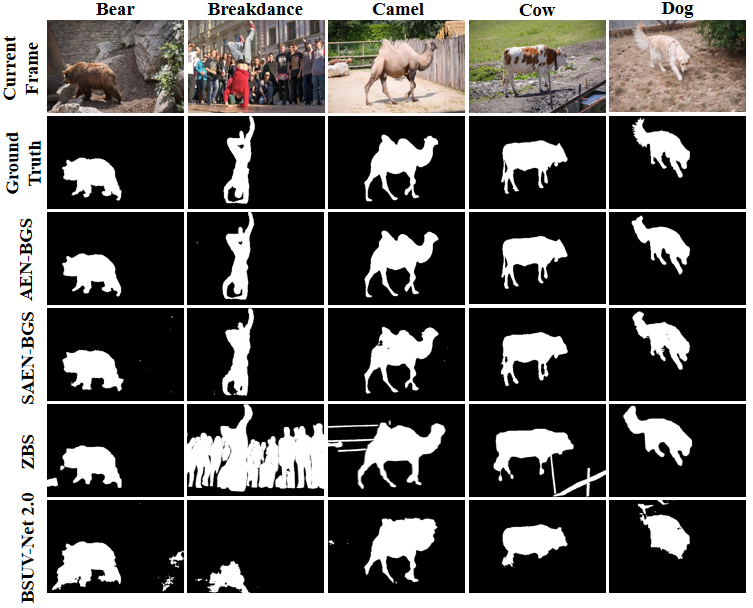}
\caption{Visual comparison of top BGS algorithms on sample frames from different categories of DAVIS-2016 dataset.
}
\label{fig: DAVIS-2016}
\end{figure}

\begin{table}[htb]
\centering
\fontsize{8}{11}\selectfont
\caption{Comparison of parameter number and energy between the counterpart AEN-BGS, ZBS and BSUV-Net 2.0. \textbf{BOLD} indicates the best.}
\resizebox{0.45\linewidth}{!}{
\begin{tabular}{c|c|c}
\hline
\textbf{Model} &\textbf{Param ($\downarrow$)} & \textbf{Energy ($\downarrow$)}        \\ \hline
{AEN-BGS}          &    \textbf{1.8M}  &    \textbf{4.10e+10}               \\ \hline
{ZBS \cite{zbs}}            & {105.4M}      & {8.23e+10}                  \\ \hline
{BSUV-Net 2.0 \cite{Tezcan2021}}            & {30.3M}      & {{1.92e+11}}                  \\ \hline
\end{tabular}}
\label{all-NN}
\end{table}

\begin{table*}[htb]
\centering
\caption{Energy performance comparison between our proposed SAEN-BGS and its counterpart AEN-BGS under the small training set (randomly selected 200 frames per video) on CDnet-2014 dataset. \textbf{BOLD} indicates the best.}
\resizebox{\linewidth}{!}{
\begin{tabular}{ccccccccccc}
\\ \hline
\multicolumn{1}{c|}{\textbf{Scene}} & \multicolumn{1}{c|}{\textbf{Model}} & \multicolumn{1}{c|}{\textbf{Rec ($\uparrow$)}} & \multicolumn{1}{c|}{\textbf{Spe ($\uparrow$)}} & \multicolumn{1}{c|}{\textbf{FPR ($\downarrow$)}}    & \multicolumn{1}{c|}{\textbf{FNR ($\downarrow$)}}    & \multicolumn{1}{c|}{\textbf{PWC ($\downarrow$)}}    & \multicolumn{1}{c|}{\textbf{Fm ($\uparrow$)}} & \multicolumn{1}{c|}{\textbf{Pre ($\uparrow$)}} & \multicolumn{1}{c|}{\textbf{AvR$_s$ ($\downarrow$)}} & \textbf{Energy ($\downarrow$)}   \\ \hline
\multirow{2}{*}{\textbf{Baseline}}   &\multicolumn{1}{|c|}{AEN-BGS}   & \multicolumn{1}{c|}{\textbf{95.05}} & \multicolumn{1}{c|}{\textbf{99.81}}      & \multicolumn{1}{c|}{\textbf{0.18}} & \multicolumn{1}{c|}{\textbf{4.94}} & \multicolumn{1}{c|}{\textbf{0.375}} & \multicolumn{1}{c|}{\textbf{95.18}}    & \multicolumn{1}{c|}{\textbf{95.31}}    & \multicolumn{1}{c|}{100.0}                  & 4.10e+10 \\ 
&\multicolumn{1}{|c|}{SAEN-BGS}  & \multicolumn{1}{c|}{94.07} & \multicolumn{1}{c|}{99.78}      & \multicolumn{1}{c|}{0.21} & \multicolumn{1}{c|}{5.92} & \multicolumn{1}{c|}{0.430} & \multicolumn{1}{c|}{94.19}    & \multicolumn{1}{c|}{94.30}    & \multicolumn{1}{c|}{\textbf{12.06}}            & \textbf{3.84e+9}  \\ \hline
\multirow{2}{*}{\textbf{Bad.Wthr}}   &\multicolumn{1}{|c|}{AEN-BGS}   & \multicolumn{1}{c|}{\textbf{93.36}} & \multicolumn{1}{c|}{\textbf{99.91}}      & \multicolumn{1}{c|}{\textbf{0.08}} & \multicolumn{1}{c|}{\textbf{6.63}} & \multicolumn{1}{c|}{\textbf{0.196}} & \multicolumn{1}{c|}{\textbf{94.30}}    & \multicolumn{1}{c|}{\textbf{95.27}}    & \multicolumn{1}{c|}{100.0}                  & 4.10e+10 \\ 
&\multicolumn{1}{|c|}{SAEN-BGS}  & \multicolumn{1}{c|}{92.32} & \multicolumn{1}{c|}{99.83}      & \multicolumn{1}{c|}{0.16} & \multicolumn{1}{c|}{7.67} & \multicolumn{1}{c|}{0.327} & \multicolumn{1}{c|}{92.37}    & \multicolumn{1}{c|}{92.42}    & \multicolumn{1}{c|}{\textbf{11.50}}            & \textbf{3.09e+9}  \\ \hline
\multirow{2}{*}{\textbf{Cam.Jtr}}   &\multicolumn{1}{|c|}{AEN-BGS}   & \multicolumn{1}{c|}{\textbf{94.29}} & \multicolumn{1}{c|}{\textbf{99.76}}      & \multicolumn{1}{c|}{\textbf{0.23}} & \multicolumn{1}{c|}{\textbf{5.70}} & \multicolumn{1}{c|}{\textbf{0.429}} & \multicolumn{1}{c|}{\textbf{93.89}}    & \multicolumn{1}{c|}{\textbf{93.48}}    & \multicolumn{1}{c|}{100.0}                  & 4.10e+10 \\ 
&\multicolumn{1}{|c|}{SAEN-BGS}  & \multicolumn{1}{c|}{90.62} & \multicolumn{1}{c|}{99.67}      & \multicolumn{1}{c|}{0.32} & \multicolumn{1}{c|}{9.30} & \multicolumn{1}{c|}{0.653} & \multicolumn{1}{c|}{91.05}    & \multicolumn{1}{c|}{91.48}    & \multicolumn{1}{c|}{\textbf{11.98}}            & \textbf{3.78e+9}  \\ \hline
\multirow{2}{*}{\textbf{Dyn.Bkgr}}   &\multicolumn{1}{|c|}{AEN-BGS}   & \multicolumn{1}{c|}{\textbf{93.95}} & \multicolumn{1}{c|}{\textbf{99.95}}      & \multicolumn{1}{c|}{\textbf{0.04}} & \multicolumn{1}{c|}{\textbf{6.04}} & \multicolumn{1}{c|}{\textbf{0.101}} & \multicolumn{1}{c|}{\textbf{94.26}}    & \multicolumn{1}{c|}{\textbf{94.58}}    & \multicolumn{1}{c|}{100.0}                  & 4.10e+10 \\ 
&\multicolumn{1}{|c|}{SAEN-BGS}  & \multicolumn{1}{c|}{91.91} & \multicolumn{1}{c|}{99.92}      & \multicolumn{1}{c|}{0.07} & \multicolumn{1}{c|}{8.08} & \multicolumn{1}{c|}{0.174} & \multicolumn{1}{c|}{92.85}    & \multicolumn{1}{c|}{93.81}    & \multicolumn{1}{c|}{\textbf{12.25}}            & \textbf{3.73e+9}  \\ \hline
\multirow{2}{*}{\textbf{Int.Obj.Mot}}   &\multicolumn{1}{|c|}{AEN-BGS}   & \multicolumn{1}{c|}{\textbf{90.11}} & \multicolumn{1}{c|}{\textbf{99.92}}      & \multicolumn{1}{c|}{\textbf{0.07}} & \multicolumn{1}{c|}{\textbf{9.88}} & \multicolumn{1}{c|}{\textbf{0.171}} & \multicolumn{1}{c|}{\textbf{91.40}}    & \multicolumn{1}{c|}{\textbf{92.73}}    & \multicolumn{1}{c|}{100.0}                  & 4.10e+10 \\ 
&\multicolumn{1}{|c|}{SAEN-BGS}  & \multicolumn{1}{c|}{86.40} & \multicolumn{1}{c|}{99.91}      & \multicolumn{1}{c|}{0.08} & \multicolumn{1}{c|}{13.59} & \multicolumn{1}{c|}{0.240} & \multicolumn{1}{c|}{89.12}    & \multicolumn{1}{c|}{92.02}    & \multicolumn{1}{c|}{\textbf{12.15}}            & \textbf{3.69e+9}  \\ \hline                  
\multirow{2}{*}{\textbf{Low.Frm.Rate}}   &\multicolumn{1}{|c|}{AEN-BGS}   & \multicolumn{1}{c|}{\textbf{93.75}} & \multicolumn{1}{c|}{\textbf{99.78}}      & \multicolumn{1}{c|}{\textbf{0.21}} & \multicolumn{1}{c|}{\textbf{6.24}} & \multicolumn{1}{c|}{\textbf{0.336}} & \multicolumn{1}{c|}{\textbf{92.03}}    & \multicolumn{1}{c|}{\textbf{90.38}}    & \multicolumn{1}{c|}{100.0}                  & 4.10e+10 \\ 
&\multicolumn{1}{|c|}{SAEN-BGS}  & \multicolumn{1}{c|}{88.73} & \multicolumn{1}{c|}{99.76}      & \multicolumn{1}{c|}{0.23} & \multicolumn{1}{c|}{11.26} & \multicolumn{1}{c|}{0.475} & \multicolumn{1}{c|}{89.12}    & \multicolumn{1}{c|}{89.52}    & \multicolumn{1}{c|}{\textbf{12.62}}            & \textbf{4.20e+9}  \\ \hline
              
\multirow{2}{*}{\textbf{Nig.Vid}}   & \multicolumn{1}{|c|}{AEN-BGS}   & \multicolumn{1}{c|}{\textbf{85.23}} & \multicolumn{1}{c|}{\textbf{99.91}}      & \multicolumn{1}{c|}{\textbf{0.08}} & \multicolumn{1}{c|}{\textbf{14.76}} & \multicolumn{1}{c|}{\textbf{0.212}} & \multicolumn{1}{c|}{\textbf{87.78}}    & \multicolumn{1}{c|}{\textbf{90.50}}    & \multicolumn{1}{c|}{100.0}                  & 4.10e+10 \\ 
&\multicolumn{1}{|c|}{SAEN-BGS}  & \multicolumn{1}{c|}{80.48} & \multicolumn{1}{c|}{99.90}      & \multicolumn{1}{c|}{0.09} & \multicolumn{1}{c|}{19.51} & \multicolumn{1}{c|}{0.265} & \multicolumn{1}{c|}{84.35}    & \multicolumn{1}{c|}{88.61}    & \multicolumn{1}{c|}{\textbf{12.46}}            & \textbf{5.09e+9}  \\ \hline

\multirow{2}{*}{\textbf{PTZ}}   & \multicolumn{1}{|c|}{AEN-BGS}   & \multicolumn{1}{c|}{\textbf{88.32}} & \multicolumn{1}{c|}{99.87}      & \multicolumn{1}{c|}{0.12} & \multicolumn{1}{c|}{\textbf{11.67}} & \multicolumn{1}{c|}{0.212} & \multicolumn{1}{c|}{\textbf{86.67}}    & \multicolumn{1}{c|}{\textbf{85.09}}    & \multicolumn{1}{c|}{100.0}                  & 4.10e+10 \\ 
&\multicolumn{1}{|c|}{SAEN-BGS}  & \multicolumn{1}{c|}{84.77} & \multicolumn{1}{c|}{\textbf{99.89}}      & \multicolumn{1}{c|}{\textbf{0.10}} & \multicolumn{1}{c|}{15.22} & \multicolumn{1}{c|}{\textbf{0.185}} & \multicolumn{1}{c|}{83.14}    & \multicolumn{1}{c|}{81.57}    & \multicolumn{1}{c|}{\textbf{12.80}}            & \textbf{4.77e+9}  \\ \hline

\multirow{2}{*}{\textbf{Shadow}}   & \multicolumn{1}{|c|}{AEN-BGS}   & \multicolumn{1}{c|}{\textbf{93.54}} & \multicolumn{1}{c|}{\textbf{99.77}}      & \multicolumn{1}{c|}{\textbf{0.22}} & \multicolumn{1}{c|}{\textbf{6.45}} & \multicolumn{1}{c|}{\textbf{0.459}} & \multicolumn{1}{c|}{\textbf{93.75}}    & \multicolumn{1}{c|}{\textbf{93.97}}    & \multicolumn{1}{c|}{100.0}                  & 4.10e+10 \\ 
&\multicolumn{1}{|c|}{SAEN-BGS}  & \multicolumn{1}{c|}{91.77} & \multicolumn{1}{c|}{99.63}      & \multicolumn{1}{c|}{0.36} & \multicolumn{1}{c|}{8.22} & \multicolumn{1}{c|}{0.647} & \multicolumn{1}{c|}{91.17}    & \multicolumn{1}{c|}{90.57}    & \multicolumn{1}{c|}{\textbf{12.38}}            & \textbf{4.47e+9}  \\ \hline
                   
\multirow{2}{*}{\textbf{Thrml}}   & \multicolumn{1}{|c|}{AEN-BGS}   & \multicolumn{1}{c|}{\textbf{94.62}} & \multicolumn{1}{c|}{\textbf{99.70}}      & \multicolumn{1}{c|}{\textbf{0.29}} & \multicolumn{1}{c|}{\textbf{5.37}} & \multicolumn{1}{c|}{\textbf{0.692}} & \multicolumn{1}{c|}{\textbf{95.48}}    & \multicolumn{1}{c|}{\textbf{96.35}}    & \multicolumn{1}{c|}{100.0}                  & 4.10e+10 \\ 
&\multicolumn{1}{|c|}{SAEN-BGS}  & \multicolumn{1}{c|}{93.39} & \multicolumn{1}{c|}{99.41}      & \multicolumn{1}{c|}{0.58} & \multicolumn{1}{c|}{6.60} & \multicolumn{1}{c|}{1.025} & \multicolumn{1}{c|}{93.02}    & \multicolumn{1}{c|}{92.66}    & \multicolumn{1}{c|}{\textbf{11.97}}            & \textbf{3.84e+9}  \\ \hline
                       
\multirow{2}{*}{\textbf{Turb}}   & \multicolumn{1}{|c|}{AEN-BGS}   & \multicolumn{1}{c|}{89.54} & \multicolumn{1}{c|}{\textbf{99.96}}      & \multicolumn{1}{c|}{\textbf{0.03}} & \multicolumn{1}{c|}{10.45} & \multicolumn{1}{c|}{\textbf{0.079}} & \multicolumn{1}{c|}{91.01}    & \multicolumn{1}{c|}{\textbf{92.54}}    & \multicolumn{1}{c|}{100.0}                  & 4.10e+10 \\ 
&\multicolumn{1}{|c|}{SAEN-BGS}  & \multicolumn{1}{c|}{\textbf{91.12}} & \multicolumn{1}{c|}{99.94}      & \multicolumn{1}{c|}{0.05} & \multicolumn{1}{c|}{\textbf{8.87}} & \multicolumn{1}{c|}{0.113} & \multicolumn{1}{c|}{\textbf{91.01}}    & \multicolumn{1}{c|}{90.90}    & \multicolumn{1}{c|}{\textbf{12.37}}            & \textbf{3.98e+9}  \\ \hline
\end{tabular}
}
\label{VAE-SVAE}
\end{table*}

\begin{table*}[htb]
\centering
\caption{Energy performance between our proposed SAEN-BGS and its counterpart AEN-BGS under the large training set (randomly selected 70\% frames per video) on CDnet-2014 dataset. \textbf{BOLD} indicates the best.}
\resizebox{\linewidth}{!}{
\begin{tabular}{ccccccccccc}
\\ \hline
\multicolumn{1}{c|}{\textbf{Scene}} & \multicolumn{1}{c|}{\textbf{Model}} & \multicolumn{1}{c|}{\textbf{Rec ($\uparrow$)}} & \multicolumn{1}{c|}{\textbf{Spe ($\uparrow$)}} & \multicolumn{1}{c|}{\textbf{FPR ($\downarrow$)}}    & \multicolumn{1}{c|}{\textbf{FNR ($\downarrow$)}}    & \multicolumn{1}{c|}{\textbf{PWC ($\downarrow$)}}    & \multicolumn{1}{c|}{\textbf{Fm ($\uparrow$)}} & \multicolumn{1}{c|}{\textbf{Pre ($\uparrow$)}} & \multicolumn{1}{c|}{\textbf{AvR$_s$ ($\downarrow$)}} & \textbf{Energy ($\downarrow$)}   \\ \hline

\multirow{2}{*}{\textbf{Baseline}}   &\multicolumn{1}{|c|}{AEN-BGS}   & \multicolumn{1}{c|}{\textbf{97.66}} & \multicolumn{1}{c|}{\textbf{99.87}}      & \multicolumn{1}{c|}{\textbf{0.12}} & \multicolumn{1}{c|}{\textbf{2.33}} & \multicolumn{1}{c|}{\textbf{0.221}} & \multicolumn{1}{c|}{\textbf{97.47}}    & \multicolumn{1}{c|}{\textbf{97.27}}    & \multicolumn{1}{c|}{100.0}                  & 4.10e+10 \\ 
&\multicolumn{1}{|c|}{SAEN-BGS}  & \multicolumn{1}{c|}{97.04} & \multicolumn{1}{c|}{99.81}      & \multicolumn{1}{c|}{0.18} & \multicolumn{1}{c|}{2.95} & \multicolumn{1}{c|}{0.308} & \multicolumn{1}{c|}{96.49}    & \multicolumn{1}{c|}{95.94}    & \multicolumn{1}{c|}{\textbf{12.42}}            & \textbf{5.52e+9}  \\ \hline

\multirow{2}{*}{\textbf{Bad.Wthr}}   &\multicolumn{1}{|c|}{AEN-BGS}   & \multicolumn{1}{c|}{\textbf{96.05}} & \multicolumn{1}{c|}{\textbf{99.96}}      & \multicolumn{1}{c|}{\textbf{0.03}} & \multicolumn{1}{c|}{\textbf{3.94}} & \multicolumn{1}{c|}{\textbf{0.089}} & \multicolumn{1}{c|}{\textbf{96.44}}    & \multicolumn{1}{c|}{\textbf{96.82}}    & \multicolumn{1}{c|}{100.0}                  & 4.10e+10 \\ 
&\multicolumn{1}{|c|}{SAEN-BGS}  & \multicolumn{1}{c|}{95.42} & \multicolumn{1}{c|}{99.94}      & \multicolumn{1}{c|}{0.05} & \multicolumn{1}{c|}{4.57} & \multicolumn{1}{c|}{0.108} & \multicolumn{1}{c|}{95.67}    & \multicolumn{1}{c|}{95.91}    & \multicolumn{1}{c|}{\textbf{12.27}}            & \textbf{5.93e+9}  \\ \hline

\multirow{2}{*}{\textbf{Cam.Jtr}}   &\multicolumn{1}{|c|}{AEN-BGS}   & \multicolumn{1}{c|}{\textbf{96.54}} & \multicolumn{1}{c|}{\textbf{99.85}}      & \multicolumn{1}{c|}{\textbf{0.14}} & \multicolumn{1}{c|}{\textbf{3.45}} & \multicolumn{1}{c|}{\textbf{0.290}} & \multicolumn{1}{c|}{\textbf{96.75}}    & \multicolumn{1}{c|}{\textbf{96.96}}    & \multicolumn{1}{c|}{100.0}                  & 4.10e+10 \\ 
&\multicolumn{1}{|c|}{SAEN-BGS}  & \multicolumn{1}{c|}{96.22} & \multicolumn{1}{c|}{99.82}      & \multicolumn{1}{c|}{0.17} & \multicolumn{1}{c|}{3.77} & \multicolumn{1}{c|}{0.337} & \multicolumn{1}{c|}{96.22}    & \multicolumn{1}{c|}{96.23}    & \multicolumn{1}{c|}{\textbf{12.41}}            & \textbf{5.43e+9}  \\ \hline

\multirow{2}{*}{\textbf{Dyn.Bkgr}}   &\multicolumn{1}{|c|}{AEN-BGS}   & \multicolumn{1}{c|}{\textbf{96.20}} & \multicolumn{1}{c|}{\textbf{99.97}}      & \multicolumn{1}{c|}{\textbf{0.02}} & \multicolumn{1}{c|}{\textbf{3.79}} & \multicolumn{1}{c|}{\textbf{0.059}} & \multicolumn{1}{c|}{\textbf{96.28}}    & \multicolumn{1}{c|}{\textbf{96.36}}    & \multicolumn{1}{c|}{100.0}                  & 4.10e+10 \\ 
&\multicolumn{1}{|c|}{SAEN-BGS}  & \multicolumn{1}{c|}{94.51} & \multicolumn{1}{c|}{99.95}      & \multicolumn{1}{c|}{0.04} & \multicolumn{1}{c|}{5.48} & \multicolumn{1}{c|}{0.084} & \multicolumn{1}{c|}{94.69}    & \multicolumn{1}{c|}{94.88}    & \multicolumn{1}{c|}{\textbf{12.77}}            & \textbf{7.49e+9}  \\ \hline

\multirow{2}{*}{\textbf{Int.Obj.Mot}}   &\multicolumn{1}{|c|}{AEN-BGS}   & \multicolumn{1}{c|}{\textbf{94.04}} & \multicolumn{1}{c|}{\textbf{99.95}}      & \multicolumn{1}{c|}{\textbf{0.04}} & \multicolumn{1}{c|}{\textbf{5.95}} & \multicolumn{1}{c|}{\textbf{0.091}} & \multicolumn{1}{c|}{\textbf{94.54}}    & \multicolumn{1}{c|}{\textbf{95.05}}    & \multicolumn{1}{c|}{100.0}                  & 4.10e+10 \\ 
&\multicolumn{1}{|c|}{SAEN-BGS}  & \multicolumn{1}{c|}{90.48} & \multicolumn{1}{c|}{99.92}      & \multicolumn{1}{c|}{0.07} & \multicolumn{1}{c|}{9.51} & \multicolumn{1}{c|}{0.158} & \multicolumn{1}{c|}{90.60}    & \multicolumn{1}{c|}{90.72}    & \multicolumn{1}{c|}{\textbf{12.69}}            & \textbf{6.30e+9}  \\ \hline

\multirow{2}{*}{\textbf{Low.Frm.Rate}}   &\multicolumn{1}{|c|}{AEN-BGS}   & \multicolumn{1}{c|}{\textbf{94.64}} & \multicolumn{1}{c|}{\textbf{99.91}}      & \multicolumn{1}{c|}{\textbf{0.08}} & \multicolumn{1}{c|}{\textbf{5.35}} & \multicolumn{1}{c|}{\textbf{0.166}} & \multicolumn{1}{c|}{\textbf{94.56}}    & \multicolumn{1}{c|}{\textbf{94.48}}    & \multicolumn{1}{c|}{100.0}                  & 4.10e+10 \\ 
&\multicolumn{1}{|c|}{SAEN-BGS}  & \multicolumn{1}{c|}{93.31} & \multicolumn{1}{c|}{99.89}      & \multicolumn{1}{c|}{0.10} & \multicolumn{1}{c|}{6.68} & \multicolumn{1}{c|}{0.209} & \multicolumn{1}{c|}{93.16}    & \multicolumn{1}{c|}{93.01}    & \multicolumn{1}{c|}{\textbf{12.91}}            & \textbf{6.05e+9}  \\ \hline

\multirow{2}{*}{\textbf{Nig.Vid}}   &\multicolumn{1}{|c|}{AEN-BGS}   & \multicolumn{1}{c|}{\textbf{88.62}} & \multicolumn{1}{c|}{\textbf{99.94}}      & \multicolumn{1}{c|}{\textbf{0.05}} & \multicolumn{1}{c|}{\textbf{11.37}} & \multicolumn{1}{c|}{\textbf{0.136}} & \multicolumn{1}{c|}{\textbf{90.75}}    & \multicolumn{1}{c|}{\textbf{92.98}}    & \multicolumn{1}{c|}{100.0}                  & 4.10e+10 \\ 
&\multicolumn{1}{|c|}{SAEN-BGS}  & \multicolumn{1}{c|}{86.50} & \multicolumn{1}{c|}{99.91}      & \multicolumn{1}{c|}{0.08} & \multicolumn{1}{c|}{13.49} & \multicolumn{1}{c|}{0.183} & \multicolumn{1}{c|}{87.67}    & \multicolumn{1}{c|}{88.86}    & \multicolumn{1}{c|}{\textbf{12.69}}            & \textbf{1.00e+10}  \\ \hline

\multirow{2}{*}{\textbf{PTZ}}   &\multicolumn{1}{|c|}{AEN-BGS}   & \multicolumn{1}{c|}{\textbf{92.58}} & \multicolumn{1}{c|}{99.95}      & \multicolumn{1}{c|}{0.04} & \multicolumn{1}{c|}{\textbf{7.41}} & \multicolumn{1}{c|}{\textbf{0.069}} & \multicolumn{1}{c|}{\textbf{87.76}}    & \multicolumn{1}{c|}{\textbf{83.41}}    & \multicolumn{1}{c|}{100.0}                  & 4.10e+10 \\ 
&\multicolumn{1}{|c|}{SAEN-BGS}  & \multicolumn{1}{c|}{89.13} & \multicolumn{1}{c|}{\textbf{99.95}}      & \multicolumn{1}{c|}{\textbf{0.04}} & \multicolumn{1}{c|}{10.86} & \multicolumn{1}{c|}{0.078} & \multicolumn{1}{c|}{85.88}    & \multicolumn{1}{c|}{82.87}    & \multicolumn{1}{c|}{\textbf{12.56}}            & \textbf{5.44e+9}  \\ \hline

\multirow{2}{*}{\textbf{Shadow}}   &\multicolumn{1}{|c|}{AEN-BGS}   & \multicolumn{1}{c|}{\textbf{95.96}} & \multicolumn{1}{c|}{\textbf{99.82}}      & \multicolumn{1}{c|}{\textbf{0.17}} & \multicolumn{1}{c|}{\textbf{4.03}} & \multicolumn{1}{c|}{\textbf{0.325}} & \multicolumn{1}{c|}{\textbf{95.83}}    & \multicolumn{1}{c|}{\textbf{95.70}}    & \multicolumn{1}{c|}{100.0}                  & 4.10e+10 \\ 
&\multicolumn{1}{|c|}{SAEN-BGS}  & \multicolumn{1}{c|}{94.24} & \multicolumn{1}{c|}{99.73}      & \multicolumn{1}{c|}{0.26} & \multicolumn{1}{c|}{5.75} & \multicolumn{1}{c|}{0.478} & \multicolumn{1}{c|}{93.89}    & \multicolumn{1}{c|}{93.54}    & \multicolumn{1}{c|}{\textbf{13.18}}            & \textbf{9.61e+9}  \\ \hline

\multirow{2}{*}{\textbf{Thrml}}   &\multicolumn{1}{|c|}{AEN-BGS}   & \multicolumn{1}{c|}{\textbf{97.25}} & \multicolumn{1}{c|}{\textbf{99.86}}      & \multicolumn{1}{c|}{\textbf{0.13}} & \multicolumn{1}{c|}{\textbf{2.74}} & \multicolumn{1}{c|}{\textbf{0.282}} & \multicolumn{1}{c|}{\textbf{97.53}}    & \multicolumn{1}{c|}{\textbf{97.81}}    & \multicolumn{1}{c|}{100.0}                  & 4.10e+10 \\ 
&\multicolumn{1}{|c|}{SAEN-BGS}  & \multicolumn{1}{c|}{94.42} & \multicolumn{1}{c|}{99.80}      & \multicolumn{1}{c|}{0.19} & \multicolumn{1}{c|}{5.57} & \multicolumn{1}{c|}{0.508} & \multicolumn{1}{c|}{95.51}    & \multicolumn{1}{c|}{96.63}    & \multicolumn{1}{c|}{\textbf{13.05}}            & \textbf{1.05e+10}  \\ \hline

\multirow{2}{*}{\textbf{Turb}}   &\multicolumn{1}{|c|}{AEN-BGS}   & \multicolumn{1}{c|}{\textbf{92.15}} & \multicolumn{1}{c|}{\textbf{99.97}}      & \multicolumn{1}{c|}{\textbf{0.02}} & \multicolumn{1}{c|}{\textbf{7.84}} & \multicolumn{1}{c|}{\textbf{0.053}} & \multicolumn{1}{c|}{\textbf{92.75}}    & \multicolumn{1}{c|}{\textbf{93.36}}    & \multicolumn{1}{c|}{100.0}                  & 4.10e+10 \\ 
&\multicolumn{1}{|c|}{SAEN-BGS}  & \multicolumn{1}{c|}{91.05} & \multicolumn{1}{c|}{99.96}      & \multicolumn{1}{c|}{0.03} & \multicolumn{1}{c|}{8.94} & \multicolumn{1}{c|}{0.064} & \multicolumn{1}{c|}{91.34}    & \multicolumn{1}{c|}{91.63}    & \multicolumn{1}{c|}{\textbf{12.42}}            & \textbf{7.29e+9}  \\ \hline
\end{tabular}
}
\label{VAE-SVAEs}
\end{table*}

\subsection{Performance Comparison}
We begin by assessing the performance on the CDnet-2014 dataset in this section. Table~\ref{table: CD2014_small} displays measurements for various algorithms when utilizing a limited training dataset. The counterpart AEN-BGS excels over the other nine methods in terms of specificity, FPR, PWC, F-measure, and precision, while also ranking second in recall and FNR. It confirms that our network designs are effective. Moreover, the proposed SAEN-BGS achieves similar performance to AEN-BGS, demonstrating that our model can still perform well with a limited training dataset. The results in Table~\ref{table: CD2014_large} for the large training set show that our SAEN-BGS approach continues to outperform other current methods. Further examination of the F-measure in 11 categories, as shown in Table~\ref{table: CD2014_per_category}, shows that our SAEN-BGS and the AEN-BGS perform the best overall, with some limitations in the shadow and baseline categories. It reflects that our design model structure can benefit to improving BGS performance. Alternatively, Fig.~\ref{fig: CD2014} provides a visual comparison of different approaches by showcasing one frame from each of seven videos spanning various categories. AEN-BGS, SAEN-BGS, and BSUV-Net 2.0 have shown superior performance compared to other baseline models. Additionally, our SAEN-BGS model shows superior visual performance compared to BSUV-Net 2.0 in categories with dynamic backgrounds such as Cam.Jir, Dyn.Bkgr, and Bad.Wthr, while achieving similar results to AEN-BGS. 

In order to assess the effectiveness of our SAEN-BGS, we compare it to two other top-performing methods (ZBS \cite{zbs} and BSUV-Net 2.0\cite{Tezcan2021}) as well as the similar AEN-BGS on the DAVIS-2016 dataset. According to Table~\ref{DAVIS-2016}, our SAEN-BGS displays a notable increase of over 15\% in F-measure compared to ZBS and BSUV-Net 2.0. Moreover, as shown in Fig. \ref{fig: DAVIS-2016}, our SAEN-BGS and the competing AEN-BGS both generate visually accurate solutions that closely match the ground truth, unlike the two other top-performing baseline methods. It is worth mentioning that DAVIS-2016 is more challenging than CDnet-2014 because it has more categories and fewer shots in each category. Hence, it is evident that the ZBS and BSUV-Net 2.0 models can easily be deceived by the intricate background objects in DAVIS-2016. Our proposed SAEN-BGS and its counterpart AEN-BGS are able to circumvent this issue.

Overall, these experiments effectively showcase our suggested techniques for improving BGS performance, as our SAEN-BGS has proven to surpass the current benchmarks. When dealing with practical situations, our main focus is on achieving efficient energy consumption, which involves striking a balance between effectiveness and energy efficiency. Ultimately, efficient energy consumption decreases the amount of active neurons, but this comes at the expense of effectiveness. We will validate the energy performance of AEN-BGS and SAEN-BGS in the upcoming experiments. 

\subsection{Energy Comparison}
We not only improve segmentation performance but also verify our power consumption. {According to Eq. \ref{flops_ann} and Eq. \ref{flops_snn}, it is evident that the calculation of flops for ANNs is independent of the particular task, unlike SNNs. In order to do this, we initially calculate the parameter count and energy of AEN-BGS, BSUV-Net 2.0, and ZBS, all of which rely on ANNs and demonstrate superior performance compared to other baseline models. According to the data in Table \ref{all-NN}, AEN-BGS has significantly fewer parameters and uses less energy than both ZBS and BSUV-Net 2.0. We will then compare our SAEN-BGS with its counterpart CNN (AEN-BGS) on CDnet-2014 and DAVIS-2016 individually to demonstrate the energy efficiency of our SAEN-BGS.}

\begin{table*}[htb]
\centering
\caption{Energy performance between our proposed SAEN-BGS and its counterpart AEN-BGS on DAVIS-2016 dataset. \textbf{BOLD} indicates the best.}
\resizebox{0.5\linewidth}{!}{
\begin{tabular}{c|c|c|c}
\hline
\textbf{Model} & \textbf{Fm ($\uparrow$)} &\textbf{AvR$_s$ ($\downarrow$)} & \textbf{Energy ($\downarrow$)}        \\ \hline
{AEN-BGS}         &  {\textbf{87.40}}         &    {100.0}  &    {4.10e+10}               \\ \hline
{SAEN-BGS}          & {85.20}           & {\textbf{13.97}}      & {\textbf{1.17e+10}}                  \\ \hline
\end{tabular}}\label{davis-energy}
\end{table*}

In CDnet-2014, we evaluate not just the seven fundamental metrics but also two additional energy metrics (Energy and AvR$_s$) across 11 categories, following the two previously mentioned training schemes. Our SAEN-BGS demonstrates similar performance to AEN-BGS across the seven fundamental metrics, as shown in Table \ref{VAE-SVAE} and Table \ref{VAE-SVAEs}. Our method achieves significantly lower AvR$_s$ and energy consumption with an average neuron activation of only 12\% and energy savings ranging from 74\% to 92\%. Furthermore, we also perform energy comparisons on DAVIS-2016 using a large training set scheme. As shown in table \ref{davis-energy}, our SAEN-BGS achieves a comparable F-measure with significantly lower energy usage, with just 13.97\% neuron activation and almost 71\% less energy consumed. 

In summary, SAEN-BGS is suggested for situations with limited energy, whereas AEN-BGS is the choice for less crucial energy limitations.

\section{Conclusion}
This paper proposes a spiking autoencoder network (termed SAEN-BGS) for background subtraction to address sensitivity to temporal background noise. The continuous spiking conv-dconv block is designed for temporal feature refinement, which achieves an average 7\% higher F-measure on CDnet-2014 and 16\% higher F-measure on average over DAVIS-2016 compared with other baselines. These results demonstrate our superior capability to disentangle foreground temporal patterns from noisy backgrounds. Then, we present the self-auxiliary spiking learning under the ANN-to-SNN framework, which facilitates our SAEN-BGS to reduce energy by more than 50\% compared to state-of-the-art baselines while maintaining at least 70\% energy saving compared to its counterpart. These findings suggest that spike-event-driven SNNs inherently suppress background noise propagation.

This approach has produced significant outcomes on commonly utilized benchmarks. Nevertheless, it does possess specific constraints. Compared to conventional neural networks, there are fewer established libraries and frameworks for working with SNNs, which can hinder development and experimentation. Furthermore, as the size of the network increases, the tradeoff between efficiency and effectiveness can become problematic, particularly with standard training algorithms using backpropagation.

\section*{CRediT authorship contribution statement}
\textbf{Zhixuan Zhang}: Conceptualization, Methodology, Software, Writing - original draft. \textbf{Qi Liu}: Supervision, Conceptualization, Writing-review \& editing, Methodology. \textbf{Xiaopeng Li}: Writing - review \& editing.

\section*{Declaration of competing interest}
The authors declare that they have no known competing financial interests or personal relations that could have appeared to influence their work reported in this paper.

\section*{Acknowledgments}
This work was supported in part by the National Natural Science Foundation of China under Grant 62202174, inpart by the GJYC program of Guangzhou under Grant 2024D01J0081, in part by The Taihu Lake Innovation Fund for the School of Future Technology of South China University of Technology under Grant 2024B105611004, and in part by the Guangdong Provincial Key Laboratory of Human Digital Twin (2022B1212010004)

\section*{Data and Code availability}
The data and code will be made available on request.

\bibliographystyle{elsarticle-num}
\bibliography{refs.bib}

\end{document}